\documentclass{article}
\usepackage[numbers,sort&compress]{natbib}
\usepackage[final]{neurips_2024}

\usepackage[utf8]{inputenc}
\usepackage[T1]{fontenc}
\usepackage{hyperref}
\usepackage{url}
\usepackage{booktabs}
\usepackage{amsfonts,amsmath,amssymb,bm}
\usepackage{nicefrac}
\usepackage{microtype}
\usepackage[table,dvipsnames,svgnames]{xcolor}
\usepackage{graphicx}
\usepackage{tabularx,array,multirow}
\usepackage{pifont}
\usepackage{xspace}
\usepackage{fontawesome5}
\usepackage{enumitem}
\setlist[itemize]{leftmargin=1.2em}
\usepackage{tikz}
\usetikzlibrary{arrows.meta,positioning,calc,fit,backgrounds}
\usepackage[font=small,labelfont=bf]{caption}
\usepackage{subcaption}

\hypersetup{colorlinks,urlcolor=purple,citecolor=cyan,linkcolor=purple}
\usepackage[capitalize]{cleveref}
\definecolor{reportblue}{HTML}{2278B5}
\definecolor{reportteal}{HTML}{2A9D8F}
\definecolor{reportorange}{HTML}{E07A2F}

\newcommand{\method}{One Shot Any View}

\newif\ifreviewmarks
\reviewmarkstrue
\ifreviewmarks
  
  \newcommand{\todo}[1]{\textcolor{orange!85!black}{[\textsc{todo:} #1]}}
  
\else
  
  \newcommand{\todo}[1]{}
  
\fi

\hypersetup{pdftitle={One Shot, Any View: High-Fidelity Video Reshooting with Geometry Guidance and Visual References},pdfauthor={Kai Huang, Runze Xu, Hui Ren, Taiyi Wu, Xinyuan Li, Xun Gao, Quande Liu, Dong Tang, Tongtong Shen, Xuefeng Su, Alan Zhao},pdfsubject={Technical report}}

\newif\ifhtmlbuild
\ifcsname lx@intercol\endcsname \htmlbuildtrue \fi
\newcommand{\projectpage}{https://khuangvision.github.io/one_shot_any_view/}
\newcommand{\projectpageline}{\textbf{Homepage:}~\href{\projectpage}{\nolinkurl{khuangvision.github.io/one_shot_any_view}}}

\title{One Shot, Any View: High-Fidelity Video Reshooting \\ with Geometry Guidance and Visual References}
\ifhtmlbuild
\author{{Kai Huang$^{*}$, Runze Xu$^{*}$, Hui Ren, Taiyi Wu, Xinyuan Li, Xun Gao\\
  Quande Liu, Dong Tang, Tongtong Shen, Xuefeng Su$^{\dagger}$, Alan Zhao$^{\dagger}$}}
\makeatletter
\lx@add@affiliation{\textbf{Online Video BU, Tencent}\\
  {\small \textsuperscript{*}Equal contribution.\quad \textsuperscript{\textdagger}Project leads.}\\
  {\small \projectpageline}}
\makeatother
\else
\author{%
  {\small Kai Huang\textsuperscript{*}, Runze Xu\textsuperscript{*}, Hui Ren, Taiyi Wu, Xinyuan Li, Xun Gao}\\[1pt]
  {\small Quande Liu, Dong Tang, Tongtong Shen, Xuefeng Su\textsuperscript{\(\dagger\)}, Alan Zhao\textsuperscript{\(\dagger\)}}\\[2pt]
  \textbf{Online Video BU, Tencent}\\[2pt]
  {\small\faGlobe~\projectpageline}}
\fi

\begin{document}
\maketitle

\ifhtmlbuild\else
\makeatletter
\let\original@makefnmark\@makefnmark
\def\@makefnmark{}
\makeatother
\footnotetext{\textsuperscript{*}Equal contribution.\quad \textsuperscript{\(\dagger\)}Project leads.}
\makeatletter
\let\@makefnmark\original@makefnmark
\makeatother
\fi

\begin{abstract}
Filming a scene from an additional angle after production usually requires a costly reshoot.
We present \method{}, a system that takes a recorded clip and generates a reshot video of the same performance along a user-specified camera trajectory.
As in recent geometry-grounded methods, the system conditions generation on the source video and an explicit 3D prior, namely a \textit{point video} rendered along the requested trajectory from a feed-forward reconstruction of the source video.
We target human-centric cinematic videos, which are often shot with long lenses, shallow depth of field, and low-key lighting, all of which degrade feed-forward reconstruction.
Existing geometry-grounded methods tend to inherit the degraded point video, and their output quality drops visibly.
Two designs make such point videos usable.
First, in addition to rendered videos of 3D assets, we train on real-world multi-camera videos, whose point videos are produced by the same reconstruction pipeline as at inference.
This training--inference consistency enables the generator to take only the composition of people and scene elements from the point video, and their appearance from the source video.
Second, a geometry-anchored RoPE offset keeps the generated video temporally synchronized with the source while largely preventing the output from copying the degraded appearance of the point video.
Meanwhile, since the generator is a pretrained reference-to-video (R2V) model, the system also accepts optional character and scene reference images, which supply the appearance of regions that the source never shows.
In both objective and human evaluations, \method{} outperforms existing reshooting methods.
\end{abstract}

\vspace{-2mm}
\begin{center}
\includegraphics[width=\dimexpr\textwidth-2\leftmargini\relax]{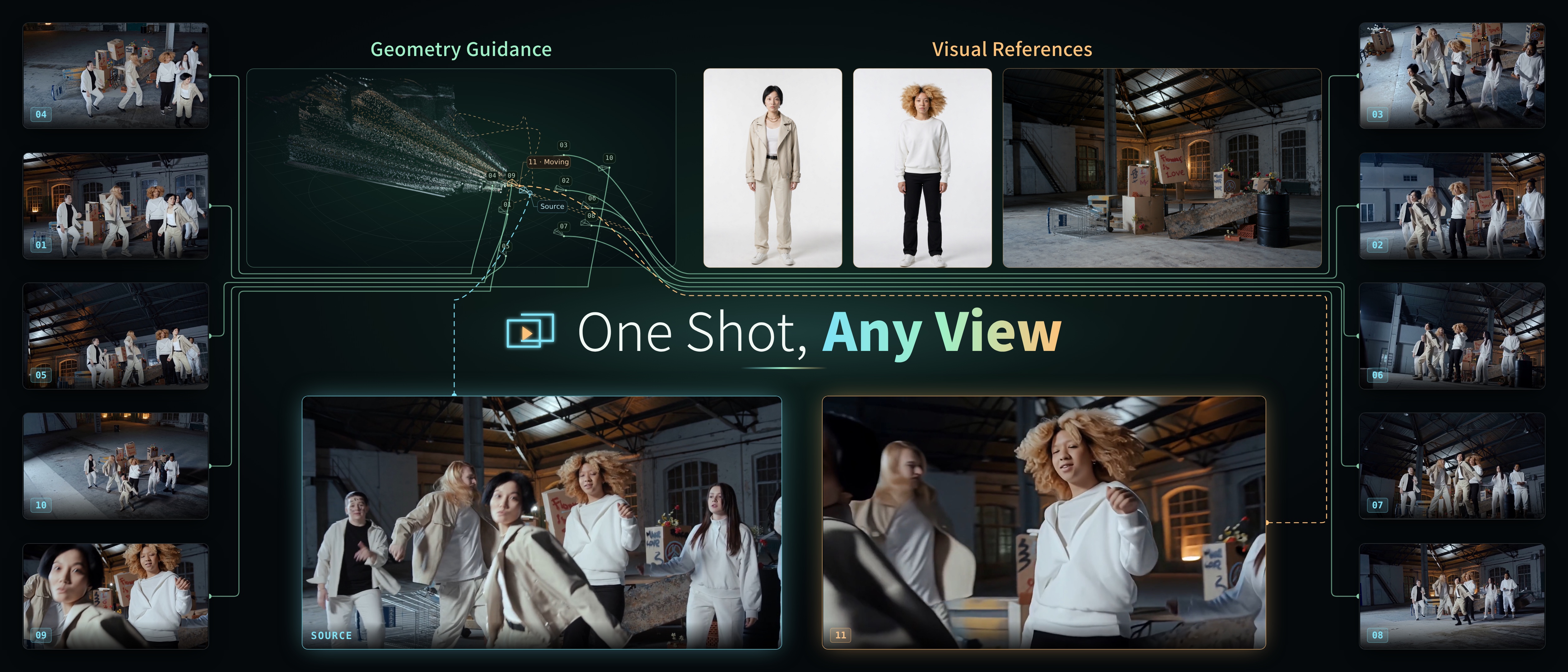}
\captionof{figure}{\textbf{One shot, any view.} Given a single source video (bottom left), \method{} uses reconstructed geometry (top left) and optional character and scene references (top right) to generate ten fixed target views (01--10, side columns) and one moving-camera view (11, bottom right). Each target video is generated independently.}
\label{fig:teaser}
\end{center}

\clearpage

\section{Introduction}
\label{sec:intro}
Camera placement decides what part of a scene is visible, how performers are framed, and how an action reads on screen. 
Once a scene has been filmed, these decisions are expensive to revise, because a different angle or an additional piece of coverage usually means bringing back the cast and crew, and rebuilding the set.
This report presents \method{}, a system that takes a single recorded shot and generates the same performance as seen from a user-specified camera trajectory.
Several target trajectories can be requested for one source video, while the reconstruction is computed once and shared, and each target video is generated independently.
Each target video changes \emph{how} the performance is observed but not the performance itself: who appears, what they wear, what they do, and when all stay the same.

\method{} follows the geometry-grounded design adopted by recent camera-controlled generation methods such as GEN3C~\citep{ren2025gen3c} and Vista4D~\citep{lin2026vista4d}.
A feed-forward 3D reconstruction model lifts the source video into a time-varying colored point cloud. Rendering it along the requested camera trajectory gives a \emph{point video}, and an R2V video diffusion model generates the target view conditioned on the source video and the point video.

We target a specific kind of input, which we call \emph{human-centric cinematic videos}: shots from films, series, and similar productions in which one or more performers occupy most of the frame.
Such shots are often filmed with long lenses and shallow depth of field.
As a result, the background, which supplies most of the stable geometric cues, is blurred.
The training data of feed-forward reconstruction models~\citep{wang2025pi3,keetha2025mapanything} consist mostly of indoor rooms, street scenes, buildings, and landscapes, and contain few close-ups of people. These models therefore degrade on such videos.
The degradation matters because it falls on the performers, on whom viewers focus most closely. A small error in gaze, expression, or hand pose is immediately noticeable.

\begin{figure}[t]
\centering\setlength{\tabcolsep}{1.2pt}\renewcommand{\arraystretch}{0.6}
\newcommand{\exrow}[1]{%
  \includegraphics[width=.318\textwidth]{figures/fig02_extreme/extreme_#1_src.jpg}&
  \includegraphics[width=.318\textwidth]{figures/fig02_extreme/extreme_#1_pc.jpg}&
  \includegraphics[width=.318\textwidth]{figures/fig02_extreme/extreme_#1_gen.jpg}}
\begin{tabular}{@{}r@{\hspace{3pt}}ccc@{}}
& \small Source video & \small Point video (target view) & \small Generated target view\\[2pt]
\rotatebox{90}{\scriptsize\quad(a) large view change} & \exrow{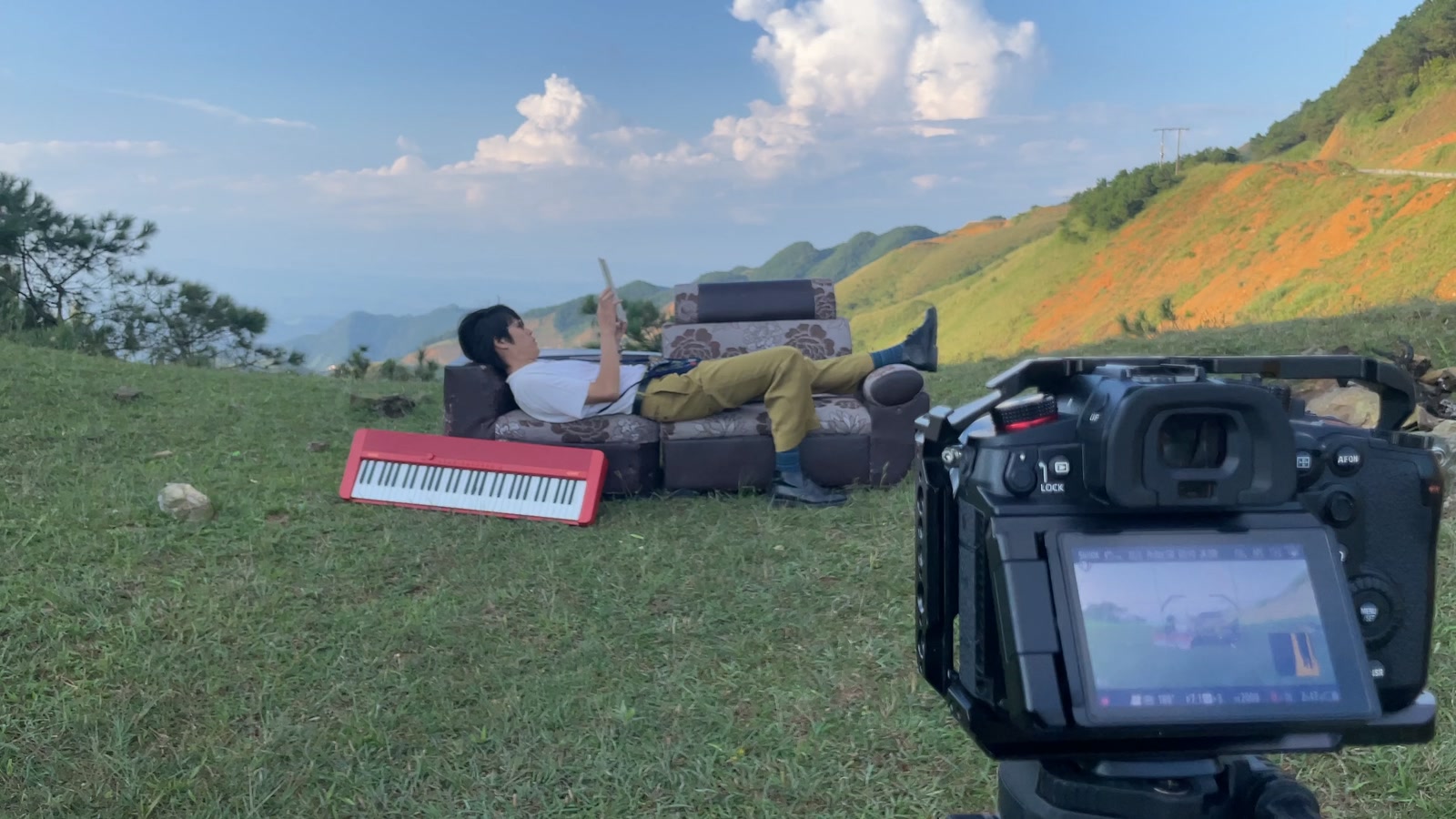}\\
\rotatebox{90}{\scriptsize\quad(b) face deformation} & \exrow{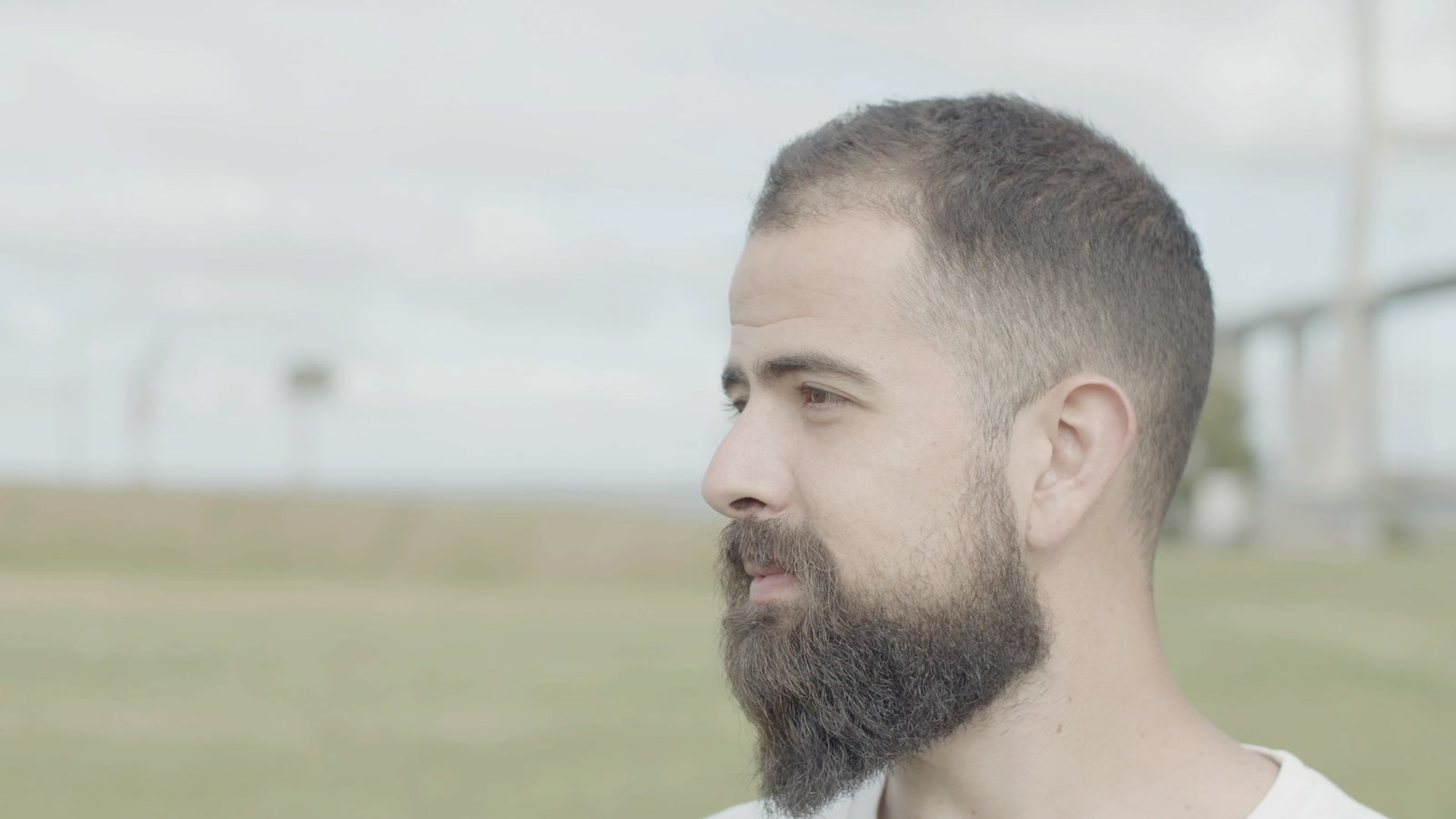}\\
\rotatebox{90}{\scriptsize\quad(c) large view change} & \exrow{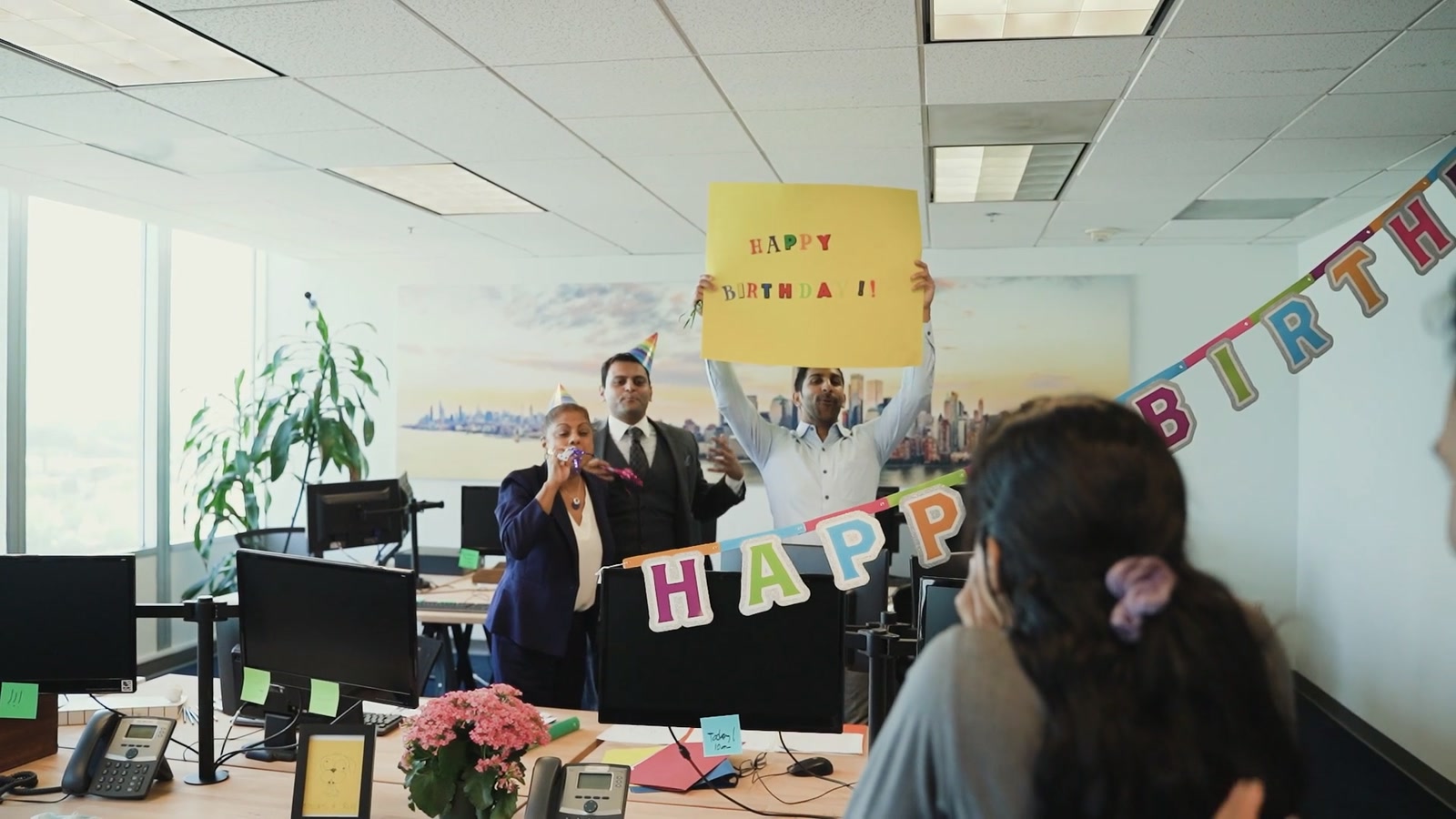}\\
\end{tabular}
\caption{\textbf{Reshooting from degraded target-view point videos.} Each row shows the source frame, the point video frame at the requested viewpoint, and the generated frame after super-resolution. In (a), the target camera looks down on a person reclining outdoors. The point video stretches the person and the seat into streaks, while the generated view keeps the person and the seat in place as seen from above and fills the holes. In (b), moving the camera from a side view toward a frontal view leaves part of the face distorted and fragmented in the point video, while the generated view completes the face with high quality. In (c), only scattered points remain after a large viewpoint change in an office scene. The generated view shows the people around the desk from the elevated target viewpoint and keeps them in their original positions.}
\label{fig:extreme}
\end{figure}

The resulting point videos show two typical failures (Figure~\ref{fig:extreme}).
First, the reconstruction captures the high-frequency depth variation of a face poorly. Once the camera moves, the face in the point video is visibly deformed (Figure~\ref{fig:extreme}b).
Second, under large viewpoint changes, the details of both the people and the scene become unrecognizable. Only their rough placement and their orientation relative to the source remain (Figure~\ref{fig:extreme}a and~c).
A practical system cannot avoid such point videos at inference. Even so, \method{} follows the requested viewpoint and completes the person and the scene consistently with the source.
We regard this robustness to degraded point videos as the most important property of the system.

We obtain this robustness mainly through a two-stage training scheme.
Rendered videos of 3D assets come with exact depth, so their point videos are clean.
Training only on these videos in the first stage quickly teaches the R2V model what a point video is and what the reshooting task asks for.
A model trained only on such data, however, learns to copy reliable point videos, and it has never seen the degraded point videos produced by real inputs.
In the second stage, we therefore train on a mixture in which real-world synchronized multi-camera videos make up about 90\% of the data and rendered videos the remaining 10\%.
The source and target videos of the real examples have the image quality and texture of real footage rather than of renders. Their point videos rendered from reconstructions of real footage degrade in the same way as at inference, while the target videos remain correct.
From these examples, the generator learns to take only the placement and orientation of people and scene elements from the point video, and to take their appearance from the source video.
In our experiments, this second stage markedly improves generation quality on real videos (Section~\ref{sec:abl-real}).

The generator must also use the point video without inheriting its low quality.
To this end, we present a geometry-anchored RoPE offset, which places the generated video at RoPE coordinates close to, but not identical with, those of the point video.
It keeps the generated actions synchronized with the source while largely preventing the degraded appearance of the point video from being copied into the output.

Moving the camera also brings into view regions that the source never shows, such as the far side of a face or background outside the original frame.
Vista4D~\citep{lin2026vista4d} introduces such information by editing geometry. It reconstructs a point cloud of an object that the user wants to add and inserts it into the source point cloud.
However, such explicit editing cannot repair the degraded points of people and objects that are already in the point cloud, which is exactly where reconstruction fails on human-centric cinematic videos.
We add references that show content unseen in the source video to the training examples. Therefore, the R2V generator learns to transfer appearance from them to regions that are not visible from the source viewpoint (Section~\ref{sec:references}).

In summary, our system provides the following capabilities.
\begin{itemize}\setlength{\itemsep}{1pt}
\item \textbf{Any view from one shot.} A single clip can be reshot along fixed cameras or keyframed trajectories, with an adjustable focal length.
\item \textbf{Robustness to degraded geometry.} The system follows the requested viewpoint and completes people and scenes even when the performer is barely recognizable in the point video, which is common when reshooting human-centric cinematic videos.
\item \textbf{Synchronized output with limited quality loss.} Generated actions stay synchronized with the source, and rendering artifacts of the point video, such as holes, streaks, and deformed faces, are largely kept out of the output.
\item \textbf{Reference-guided completion.} Optional character and scene references supply the appearance of regions that the source never shows.
\item \textbf{Delivery at production resolution.} The native 768P output can be further enhanced to 1440P by diffusion super-resolution.
\end{itemize}

The rest of this report is organized as follows. Section~\ref{sec:related} reviews related work. Section~\ref{sec:data} describes the data, and Section~\ref{sec:method} describes the model and how it is trained and used at inference. Section~\ref{sec:evaluation} compares the performance of \method{} with other camera-controllable generation methods, and further evaluates core designs of our system.

\section{Related Work}
\label{sec:related}
\subsection{Camera-controlled video generation}
Camera-controlled video generation lets the user specify the camera path of the output. When the input is an existing video, as in reshooting, the output must also keep the appearance and dynamics of the recorded scene.
We group these methods by whether a dense geometry signal serves as a condition to the generator.

\paragraph{Implicit geometry.}
Methods in this group encode the camera, typically as extrinsics and, where supported, intrinsics, into features that guide the generator.
The generator learns the mapping from these features to the appearance of the target view without rendering a reconstruction.
ReCamMaster conditions video re-rendering on camera trajectories~\citep{bai2025recammaster}.
CameraAnything extends camera control to joint intrinsic and extrinsic changes through ray-based conditioning~\citep{li2026cameraanything}.
CamCloneMaster takes the camera motion from a reference video instead of from explicit camera parameters~\citep{luo2025camclonemaster}.
We compare with CameraAnything as a representative of this group (Section~\ref{sec:evaluation}).

\paragraph{Explicit geometric priors.}
An alternative is to reconstruct the scene and express the requested viewpoint through a rendering of the reconstruction.
GEN3C conditions video generation on renderings of a point-cloud cache~\citep{ren2025gen3c}.
Vista4D grounds video reshooting in a 4D point cloud of the source scene and feeds the source video together with the target-view point video to a video diffusion model~\citep{lin2026vista4d}.
Holo4D projects persistent reconstructed 3D trajectories into the requested views~\citep{lan2026holo4d}, and NeoVerse renders reconstructed 4D Gaussians as novel-view conditions~\citep{yang2026neoverse}.
Such priors expose projected structure and visibility to the generator, which completes the regions that the source does not observe. Their outputs, however, depend on the quality of the reconstruction, which degrades on human-centric cinematic videos (Section~\ref{sec:intro}).
Our method follows this line and is closest in structure to Vista4D.

\subsection{Reference-to-video generation}
Text-to-video, image-to-video, and first-last-frame generation organize synthesis around a prompt or designated boundary frames.
Reference-to-video (R2V) models instead take user-supplied references, such as images, videos, or audio, and use them to specify identity, appearance, action, or scene content.
Commercial systems such as Seedance~2.5~\citep{bytedance2026seedance25}, Wan~3.0~\citep{alibaba2026wan30}, and Kling~4.0~\citep{kuaishou2026kling40} illustrate the shift toward multimodal references and editing.
Openly released models and frameworks include MiniMax~H3~\citep{minimax2026h3}, Bernini~\citep{liu2026bernini}, OmniWeaving~\citep{pan2026omniweaving}, and UniVideo~\citep{wei2025univideo}. These models differ in the modalities they support and in how semantic planning is coupled with pixel generation.
We build on an R2V foundation model, so that character and scene reference images can be supplied alongside the point video.

\section{Data}
\label{sec:data}
\label{sec:data-all}
Each training example combines a source video, a point video rendered into the target view, the recording of the target camera (the target recording), optional reference images, and a structured caption (Section~\ref{sec:data-example}).
We build such examples from two sources. Table~\ref{tab:consistency} compares them with the input that the generator receives at inference.
Videos rendered by a 3D engine come with exact geometry (Section~\ref{sec:data-rendered}). Their point videos are clean, and their appearance is rendered rather than real.
Real-world synchronized multi-camera videos come without geometry, so we estimate it with a feed-forward reconstruction model (Section~\ref{sec:data-real}). Their source videos and target recordings are real footage, and their point videos are degraded in the same way as at inference.
Reference images and structured captions are attached to examples from both sources (Sections~\ref{sec:references} and~\ref{sec:caption}).

\begin{table}[t]
\centering\small
\captionsetup{position=top,skip=18pt}
\caption{\textbf{Training data and inference input compared.} Estimated geometry comes from a feed-forward reconstruction model. Degraded point videos show deformed faces, lost detail, holes, and flicker (Figure~\ref{fig:extreme}), whereas rendered point videos have holes only where the source does not see the scene.}
\label{tab:consistency}
\begin{tabular}{@{}lccc@{}}
\toprule
& Rendered data & Real data & Inference input \\
\midrule
Training stage & 1 and 2 & 2 & -- \\
Geometry & ground truth & estimated & estimated \\
Image appearance & rendered & real & real \\
Point video quality & clean & degraded & degraded \\
\bottomrule
\end{tabular}
\end{table}

\subsection{Training example}
\label{sec:data-example}
A training example is a tuple $(V_s,G,V_t,R,S)$, where $V_s$ is the source video, $G$ the point video, $V_t$ the target recording, $R$ the optional reference images, and $S$ the structured caption (Sections~\ref{sec:references} and~\ref{sec:caption}).
$V_s$ and $V_t$ depict the same action from two synchronized cameras.
The point video $G$ is rendered frame by frame.
For every frame, each source pixel is lifted to 3D with its depth and the source camera of that frame and projected, with its color, into the target camera of the same frame. Rasterizing these points gives one frame of $G$.
$G$ therefore shares time and viewpoint with $V_t$, while its content comes from $V_s$.
Each frame is rendered from its own geometry, without accumulating points over time.
The same rendering is used for both training sources and for inference.

\subsection{Rendered data from a 3D engine}
\label{sec:data-rendered}
The rendered data are produced with a 3D engine from 3D assets.
Scenes contain animated characters and are captured by several synchronized virtual cameras.
The engine provides intrinsics, extrinsics, and ground-truth depth for every camera and frame.
For a synchronized source--target camera pair, the point video is rendered from this geometry as described in Section~\ref{sec:data-example}, and the video of the target camera serves as the target recording.
In total, the rendered data contain about 90k such source--target pairs.

\begin{figure}[tb]
\centering\setlength{\tabcolsep}{1pt}\renewcommand{\arraystretch}{0.6}
\begin{tabular}{@{}ccc@{}}
\small Source & \small Point video (target view) & \small Target recording\\[2pt]
\includegraphics[width=.325\textwidth]{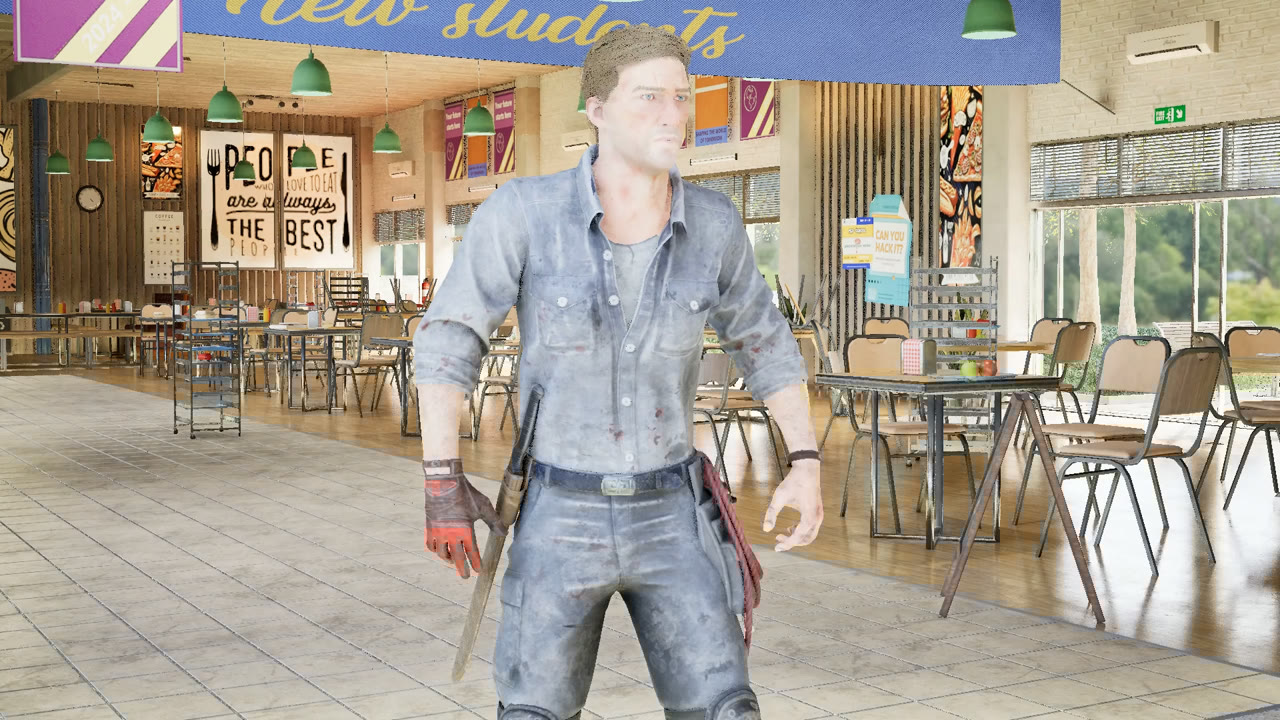}&
\includegraphics[width=.325\textwidth]{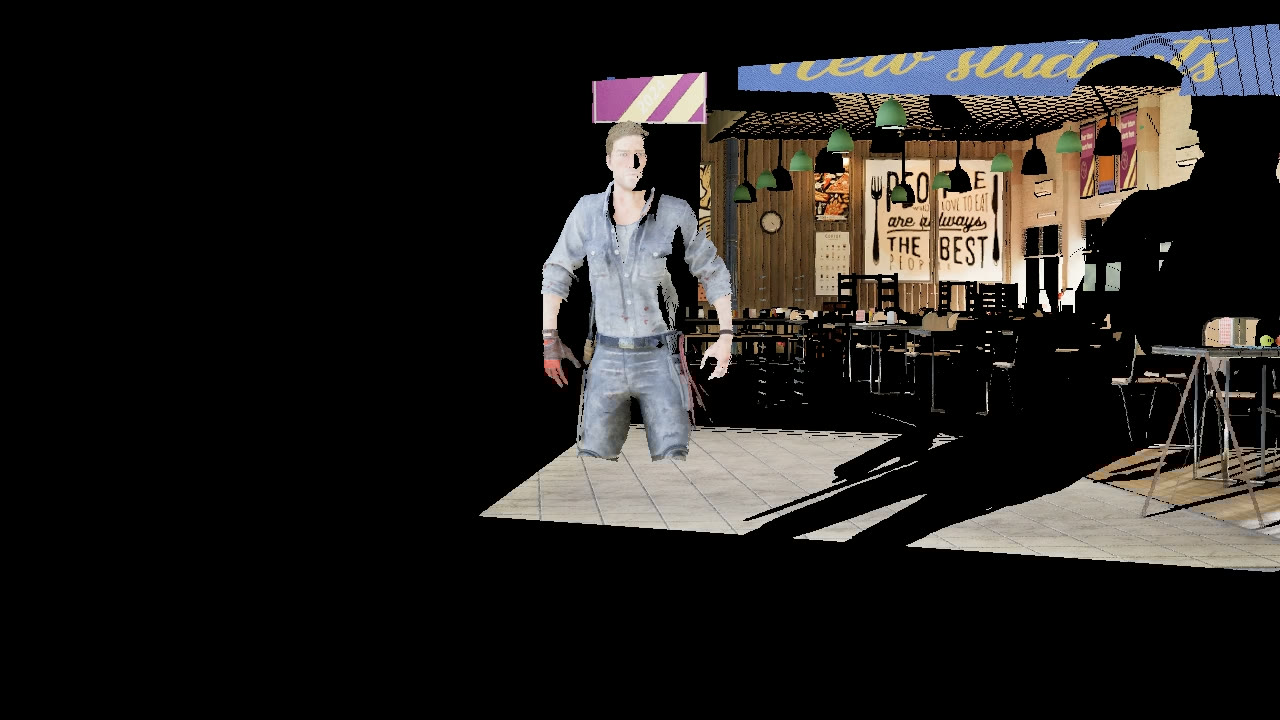}&
\includegraphics[width=.325\textwidth]{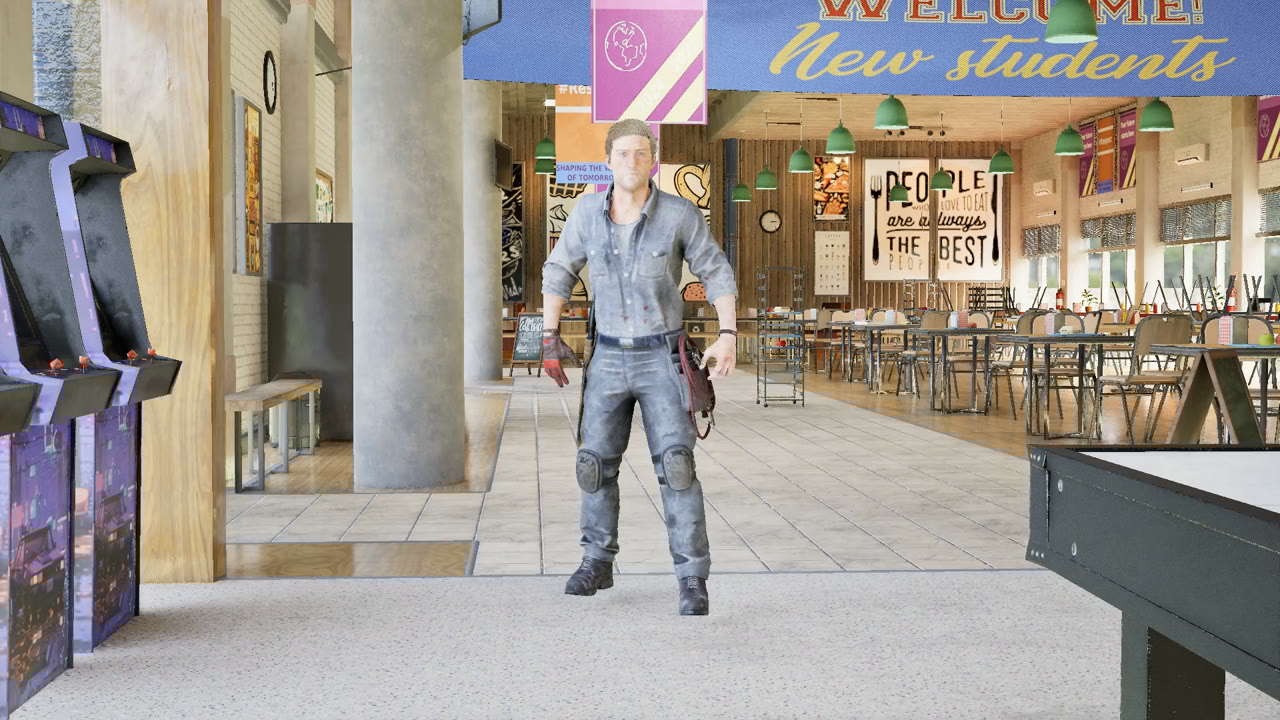}\\
\includegraphics[width=.325\textwidth]{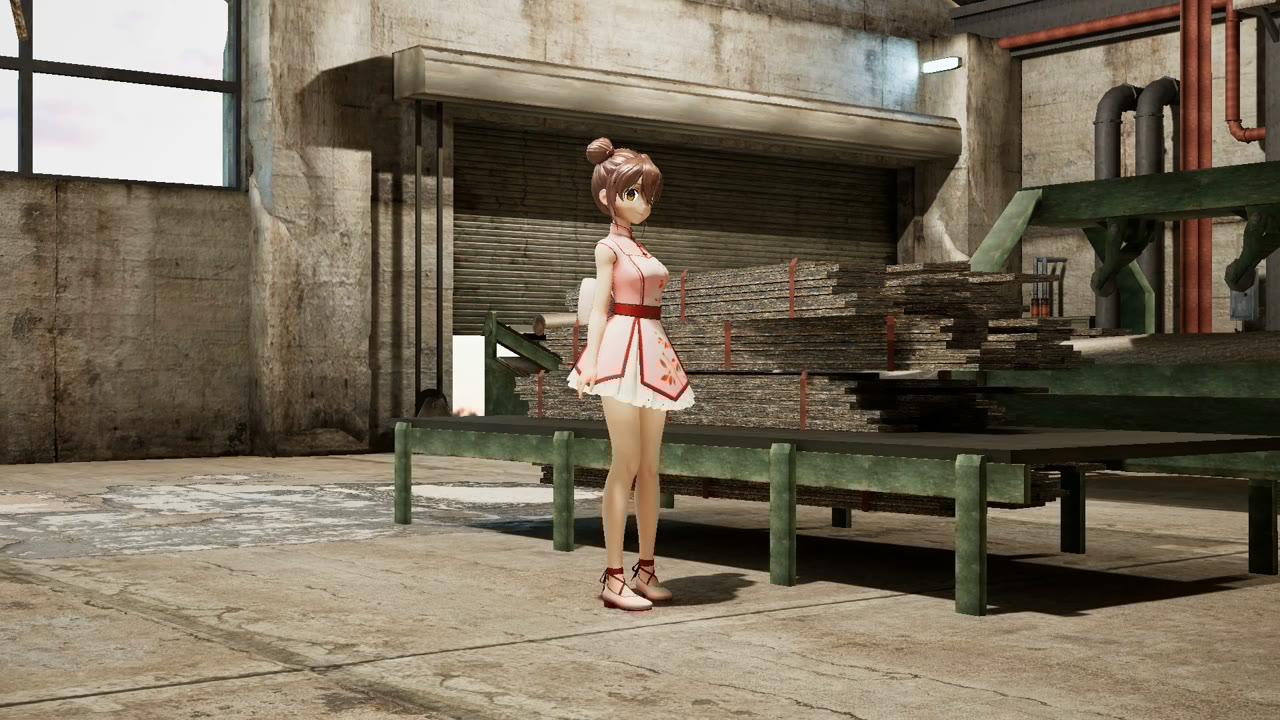}&
\includegraphics[width=.325\textwidth]{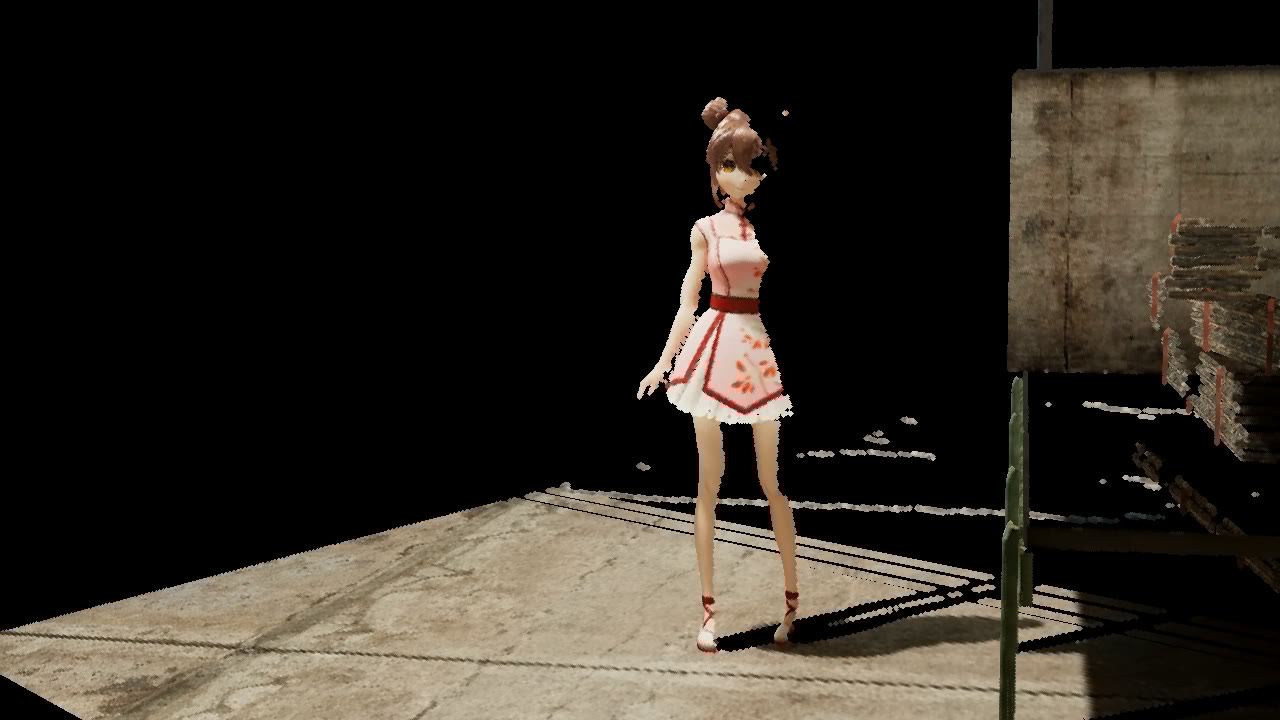}&
\includegraphics[width=.325\textwidth]{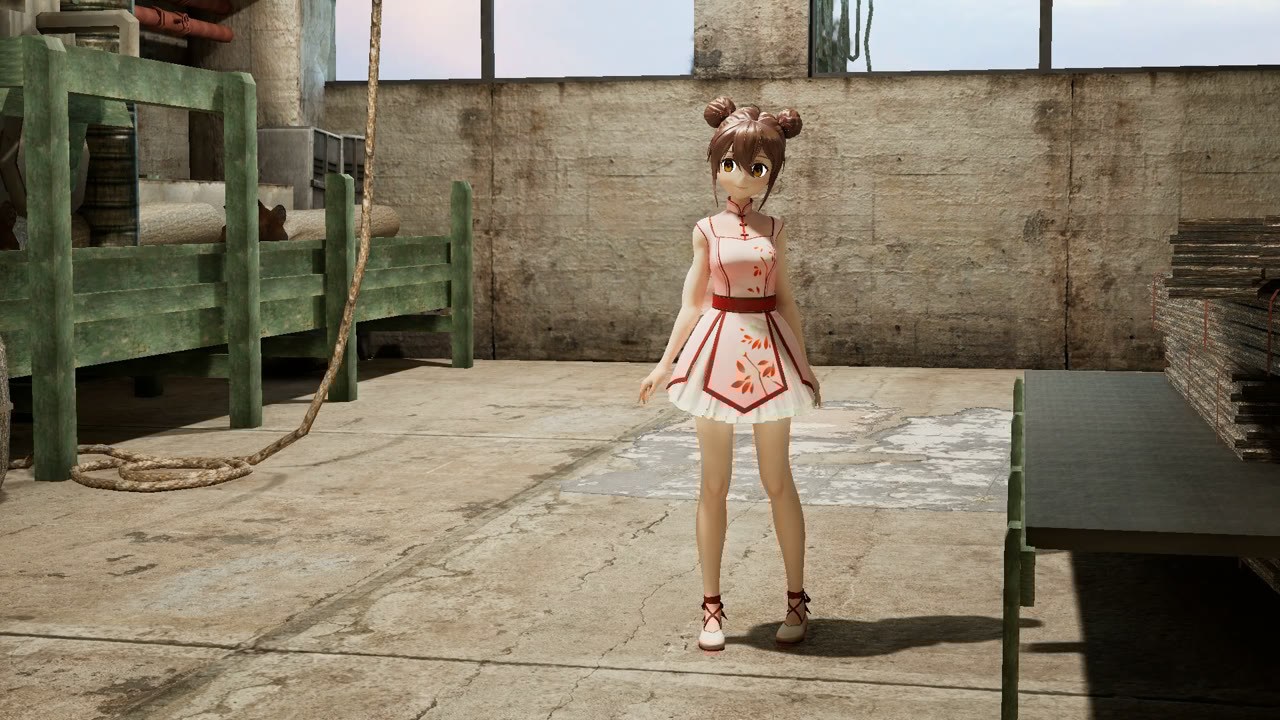}\\
\end{tabular}
\caption{\textbf{Rendered data.} Each row shows a source frame, the point-video frame rendered into the target camera from the depth of the 3D engine, and the target frame. Wherever the source sees a surface, the point video is complete and aligned with the target. Black regions appear only where the source cannot see the scene.}
\label{fig:data-rendered}
\end{figure}

Because depth and cameras are exact, the resulting point videos are markedly cleaner than those obtained from real videos (Figures~\ref{fig:data-rendered} and~\ref{fig:data-real}).
Surfaces are complete wherever the source observes them, and there is no depth noise, scale drift, or temporal jitter.
The only missing regions are disocclusions and regions outside the source frustum.
These examples therefore teach the core skill of following a target viewpoint and completing disoccluded regions, but they never show a point video that places content incorrectly.

\subsection{Real-world synchronized multi-camera data}
\label{sec:data-real}
The real data are in-house multi-camera videos in a cinematic style.
Before selection, they comprise about 30k synchronized takes, each filmed by two to four cameras.
After selection, about 10k two-camera clips of 3--15\,s remain.

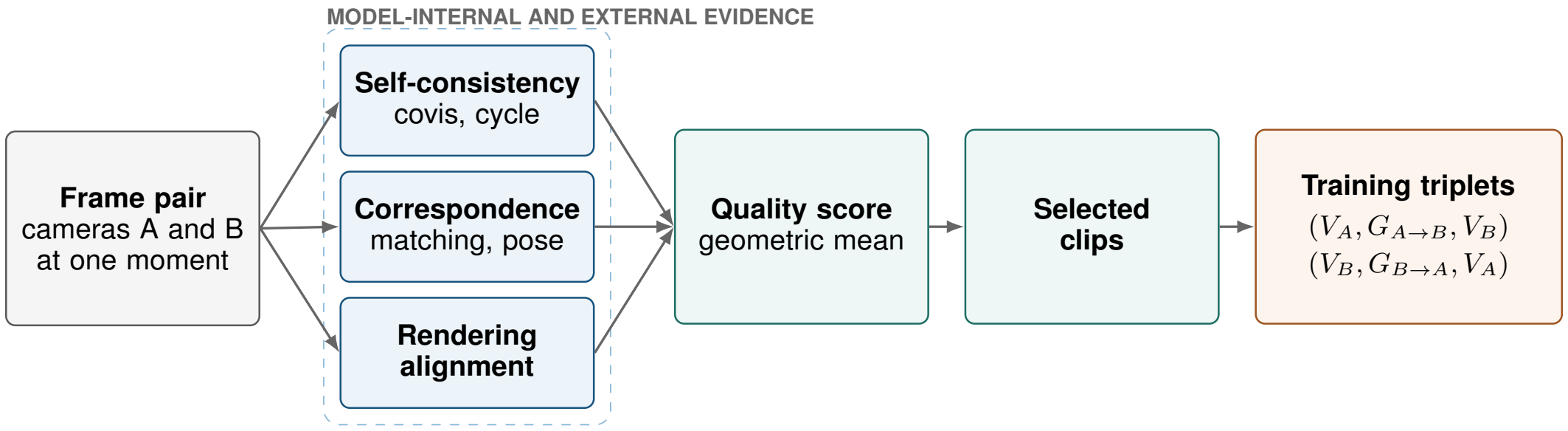
\begin{figure}[tb]
\centering
\resizebox{\textwidth}{!}{%
\begin{tikzpicture}[
  font=\sffamily\small,
  box/.style={draw=#1!70!black, fill=#1!8, rounded corners=3pt, align=center,
              minimum height=1.25cm, text width=2.7cm, inner sep=2.5pt, line width=.6pt},
  box/.default=reportblue,
  arr/.style={-{Latex[length=2mm]}, line width=.8pt, draw=black!60},
  lab/.style={font=\sffamily\scriptsize\bfseries, text=black!60, anchor=south west, inner sep=1pt}]
\node[box=gray, minimum height=2.2cm] (in) {\textbf{Frame pair}\\cameras A and B\\at one moment};
\node[box, right=9mm of in, yshift=14.5mm] (m1) {\textbf{Self-consistency}\\covis, cycle};
\node[box, below=1.6mm of m1]              (m2) {\textbf{Correspondence}\\matching, pose};
\node[box, below=1.6mm of m2]              (m3) {\textbf{Rendering}\\\textbf{alignment}};
\node[box=reportteal, right=9mm of m2, minimum height=2.2cm] (agg) {\textbf{Quality score}\\geometric mean};
\node[box=reportteal, right=4mm of agg, minimum height=2.2cm] (sel) {\textbf{Selected}\\\textbf{clips}};
\node[box=reportorange, right=4mm of sel, text width=3.3cm, minimum height=2.2cm] (out) {\textbf{Training triplets}\\[3pt]$(V_A,G_{A\to B},V_B)$\\[2pt]$(V_B,G_{B\to A},V_A)$};
\foreach \m in {m1,m2,m3} {\draw[arr] (in.east) -- (\m.west); \draw[arr] (\m.east) -- (agg.west);}
\foreach \a/\b in {agg/sel,sel/out} \draw[arr] (\a) -- (\b);
\begin{scope}[on background layer]
\node[draw=reportblue!55, dashed, rounded corners=5pt, inner sep=1.8mm, fit=(m1)(m3)] (mx) {};
\end{scope}
\node[lab] at (mx.north west) {MODEL-INTERNAL AND EXTERNAL EVIDENCE};
\end{tikzpicture}}
\caption{\textbf{Selecting real multi-camera data.} The reconstruction of each frame pair is scored by its self-consistency, by its agreement with correspondences between the real images, and by the alignment of its rendering with the target frame. For each selected clip, we render the reconstruction of camera $A$ into camera $B$ at the original frame rate to obtain $G_{A\to B}$, and swap the cameras to obtain $G_{B\to A}$.}
\label{fig:realpipe}
\end{figure}

Real recordings come with no ground-truth depth or camera parameters.
The depth and cameras needed to render a point video must therefore be estimated from the footage itself. Reliable 3D annotation of multi-camera footage is not available.
Two problems arise in our setting.
First, many camera pairs film no common content.
For example, one camera may film character A while the other films character B. A feed-forward reconstruction model still returns depth and cameras for such a pair, without any sign that the result is meaningless.
Training on such pairs is harmful.
The target cannot be predicted from the source, so the generator learns to guess.
Such pairs also never occur at inference, where the target view always shows the scene of the source.
Second, even when two cameras film the same scene, the estimated depth and camera poses can be inaccurate.
The point video then places content where the target recording does not show it, so the condition and the supervision disagree.
Checking a reconstruction against the target recording is itself difficult, because the cameras differ in viewpoint, focal length, framing, exposure, and color. Comparing the pixel colors of a rendering with the target recording is therefore unreliable.
We call the two frames of a camera pair $(A,B)$ at one sampled moment a \emph{frame pair}, score the reconstruction of every frame pair with three groups of metrics, and keep only the frame pairs that pass (Figure~\ref{fig:realpipe}).

\paragraph{Self-consistency.}
The two self-consistency scores, covis (co-visibility) and cycle, follow the geometric consistency check that multi-view stereo uses to filter depth maps~\citep{yao2018mvsnet}.
Let $\pi_{A\to B}(u)$ be the pixel of $B$ onto which pixel $u$ of $A$ projects after being lifted with its depth $D_A(u)$, and let $z_{A\to B}(u)$ be its depth in the camera of $B$. Over the pixels $\Omega_A$ that project in front of $B$ and inside its image,
\begin{align}
\mathrm{covis}&=\frac{1}{|\Omega_A|}\sum_{u\in\Omega_A}\mathbf 1\Bigl[\bigl|z_{A\to B}(u)-D_B\bigl(\pi_{A\to B}(u)\bigr)\bigr|<\epsilon\,z_{A\to B}(u)\Bigr],\\
\mathrm{cycle}&=\frac{1}{|\Omega_A|}\sum_{u\in\Omega_A}\mathbf 1\Bigl[\bigl\|\pi_{B\to A}\bigl(\pi_{A\to B}(u)\bigr)-u\bigr\|<\tau\Bigr].
\end{align}
Covis checks whether the two depth maps agree where they overlap, and cycle checks whether a pixel returns to its origin after projecting to $B$ and back with the depth of $B$.
Self-consistency rejects frame pairs whose two depth maps contradict each other, which is typical when the cameras film different content.
It compares the reconstruction only with itself, however, so it cannot detect geometry that is consistent but wrong.

\paragraph{Correspondence.}
Two-view geometry from image correspondences provides evidence independent of the reconstruction model.
An off-the-shelf dense matcher, RoMa~v2~\citep{edstedt2025romav2}, finds correspondences $\{(a_i,b_i)\}_{i=1}^{n}$ between the two real images.
The \emph{matching} score is the fraction of points $a_i$ that, projected with the reconstructed geometry, land within $\tau$ of their partners $b_i$,
\begin{equation}
s_\mathrm{match}=\frac{1}{n}\sum_{i=1}^{n}\mathbf 1\bigl[\|\pi_{A\to B}(a_i)-b_i\|<\tau\bigr].
\end{equation}
The score is scaled down when the correspondences are few or cover only a small region, which suppresses frame pairs that share little content.
For the \emph{pose} score, we estimate the relative camera pose from the correspondences with PoseLib~\citep{larsson2020poselib}, taking only the intrinsics from the reconstruction, and compare it with the relative pose of the reconstruction,
\begin{equation}
s_\mathrm{pose}=\tfrac12\bigl(2^{-\Delta_R/\theta_R}+2^{-\Delta_t/\theta_t}\bigr),
\end{equation}
where $\Delta_R$ is the angle between the two relative rotations, $\Delta_t$ is the angle between the two translation directions, and $\theta_R$ and $\theta_t$ are the angles at which the corresponding terms halve.
Matching detects depth and cameras that are self-consistent but disagree with the images.
Pose isolates errors in the estimated cameras, which shift the entire rendering even when each depth map looks plausible on its own.

\paragraph{Rendering alignment.}
This metric directly measures how well the point video aligns with the target recording, which is what training depends on.
We sample small patches $P_j$ at textured positions $p_j$ of the rendering of $A$ into $B$. Each patch is searched for in the target frame $I_B$ within a neighborhood $\mathcal N(p_j)$ by normalized cross-correlation (NCC), and the score counts patches whose best match lies close to where they were rendered,
\begin{equation}
d_j=\Bigl\|\mathop{\arg\max}_{x\in\mathcal N(p_j)}\mathrm{NCC}\bigl(P_j,I_B(x)\bigr)-p_j\Bigr\|,\qquad
s_\mathrm{render}=\frac{1}{m}\sum_{j=1}^{m}\mathbf 1[d_j<\tau].
\end{equation}
NCC compares structure rather than absolute color, so the metric is insensitive to the exposure and color differences between cameras noted above.

Every score that depends on a threshold ($\epsilon$ or $\tau$) is averaged over several thresholds that double from fine to coarse, and every score is averaged over the two directions, $A\to B$ and $B\to A$.
The five scores are then combined by a geometric mean,
\begin{equation}
S=\bigl(\mathrm{covis}\cdot\mathrm{cycle}\cdot s_\mathrm{match}\cdot s_\mathrm{pose}\cdot s_\mathrm{render}\bigr)^{1/5},
\end{equation}
so that a near-zero score on any of them pulls the total down regardless of the others.
A frame pair is accepted when $S$ passes a threshold and when the person regions of the two cameras overlap sufficiently after projection, which ensures that the two views share the foreground and the scene.
Consecutive accepted frame pairs of a camera pair form clips of 3--15\,s.
A clip is kept only if a person is visible for a sufficient share of its duration in each camera. Clips of a nearly static background would degrade the training of a video model.

\begin{figure}[tb]
\centering\setlength{\tabcolsep}{1pt}\renewcommand{\arraystretch}{0.6}
\begin{tabular}{@{}ccc@{}}
\small Source & \small Point video (target view) & \small Target recording\\[2pt]
\includegraphics[width=.325\textwidth]{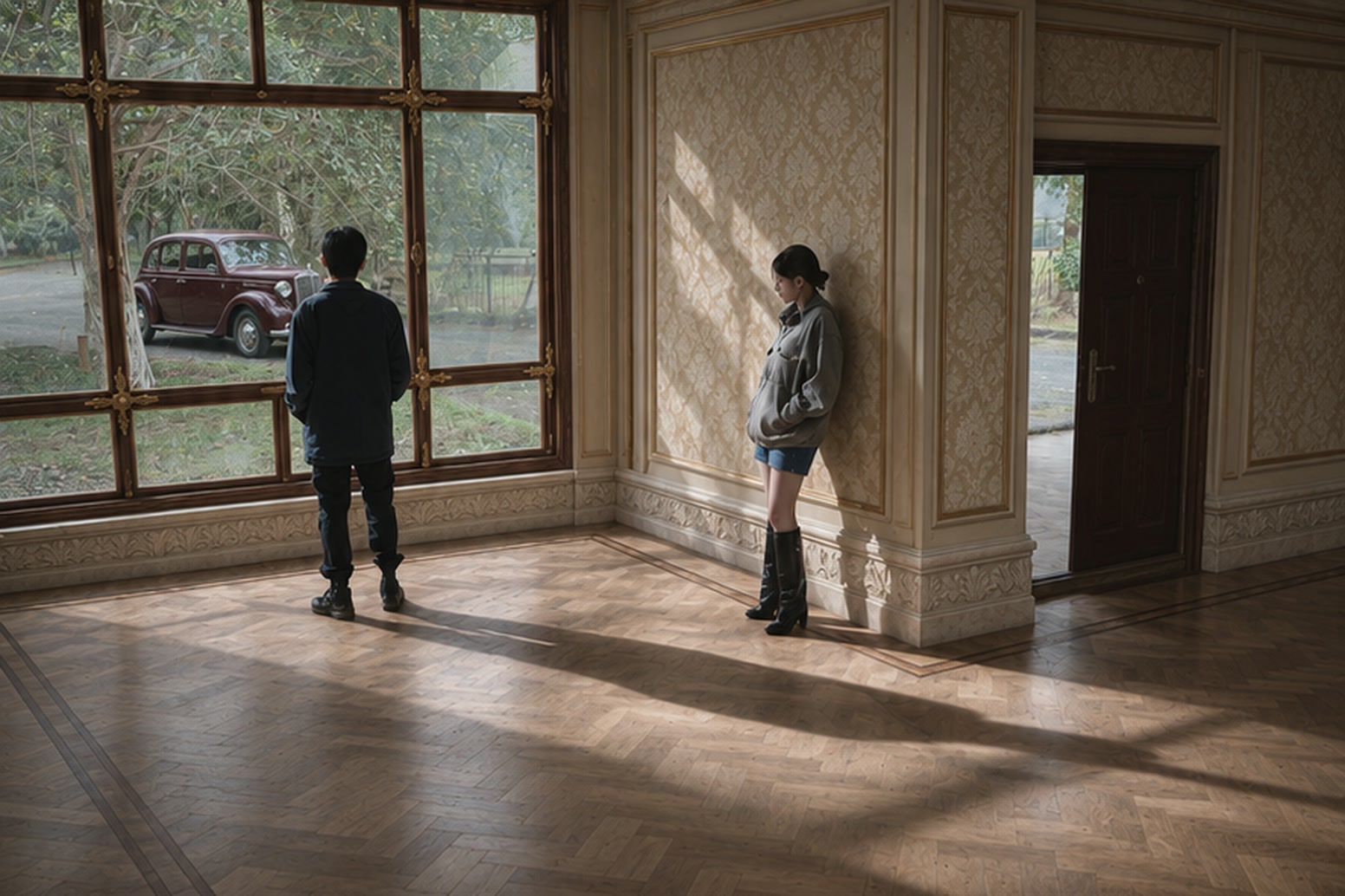}&
\includegraphics[width=.325\textwidth]{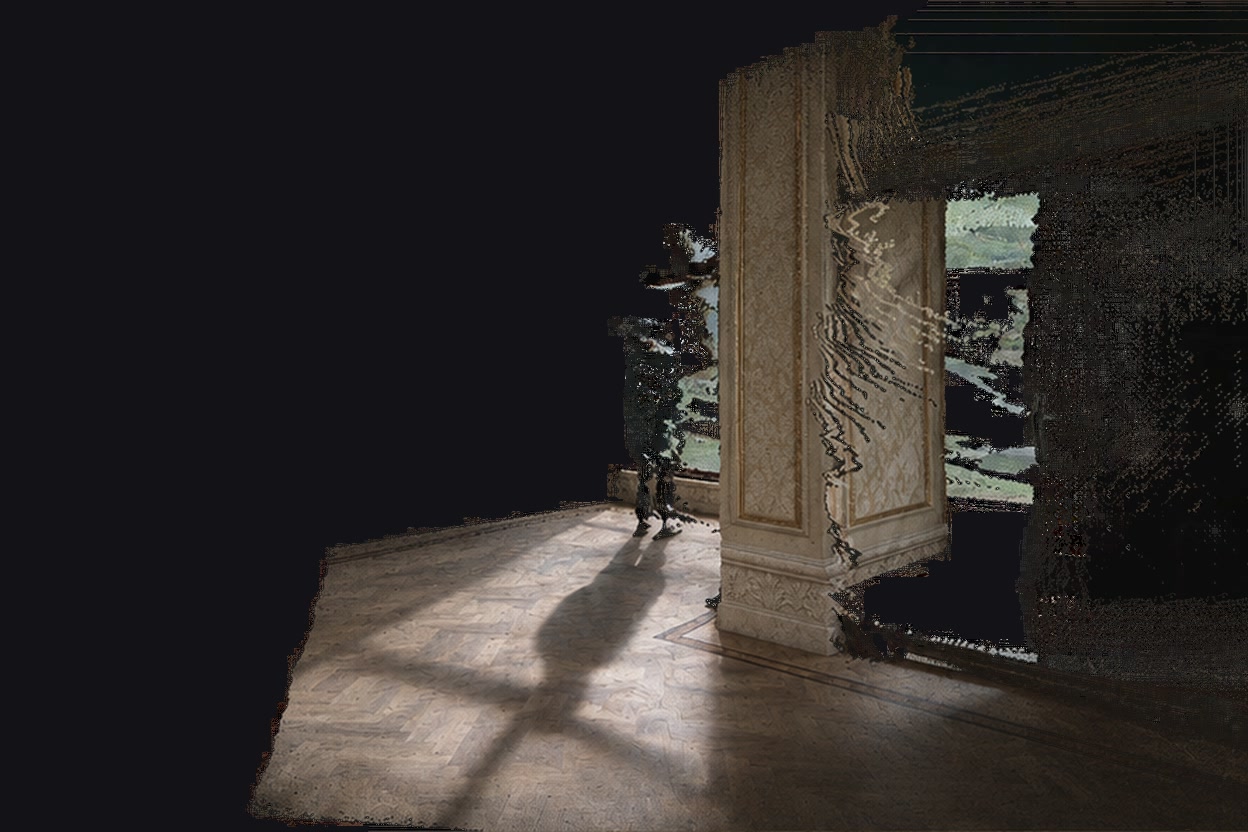}&
\includegraphics[width=.325\textwidth]{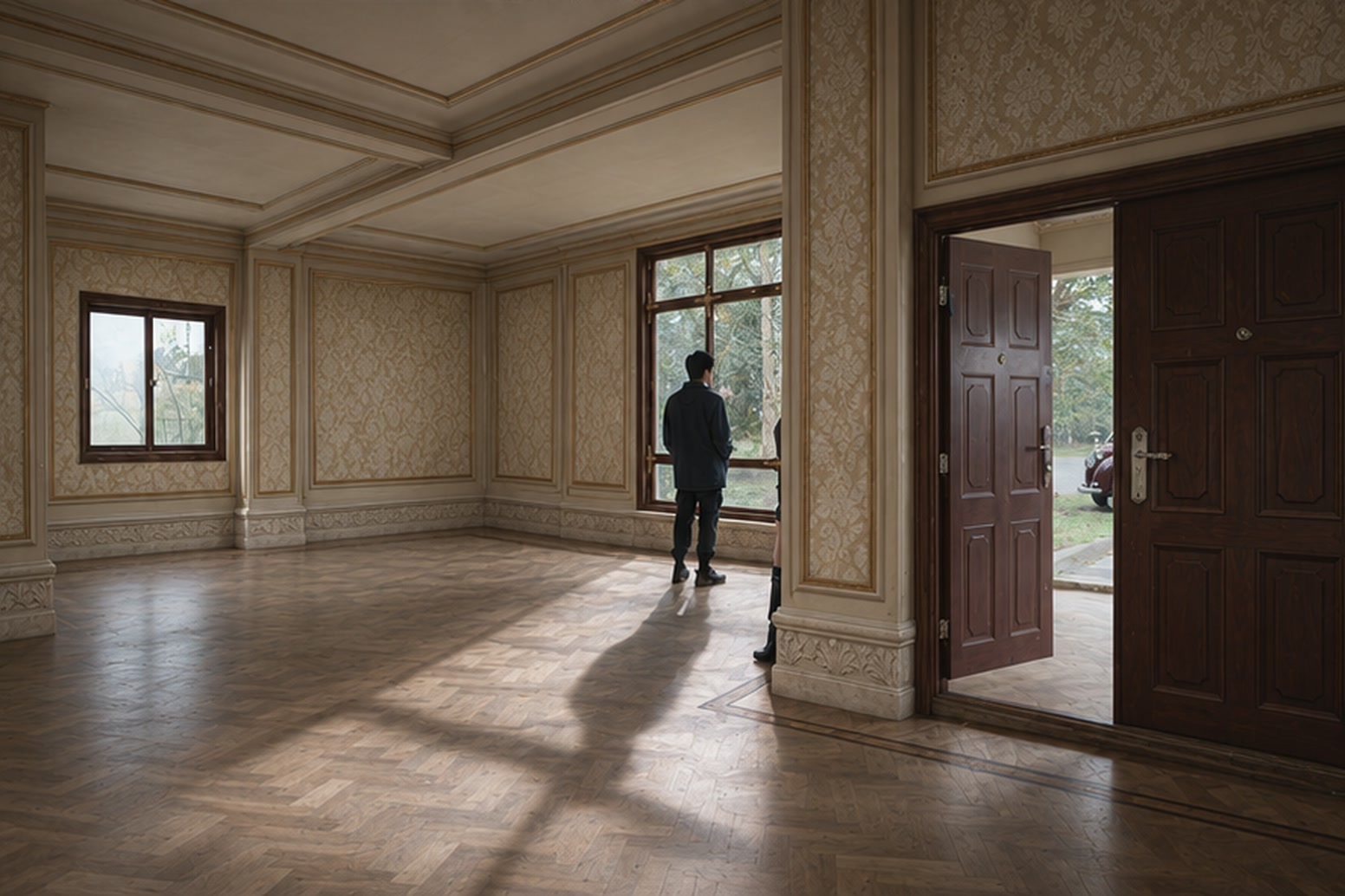}\\
\includegraphics[width=.325\textwidth]{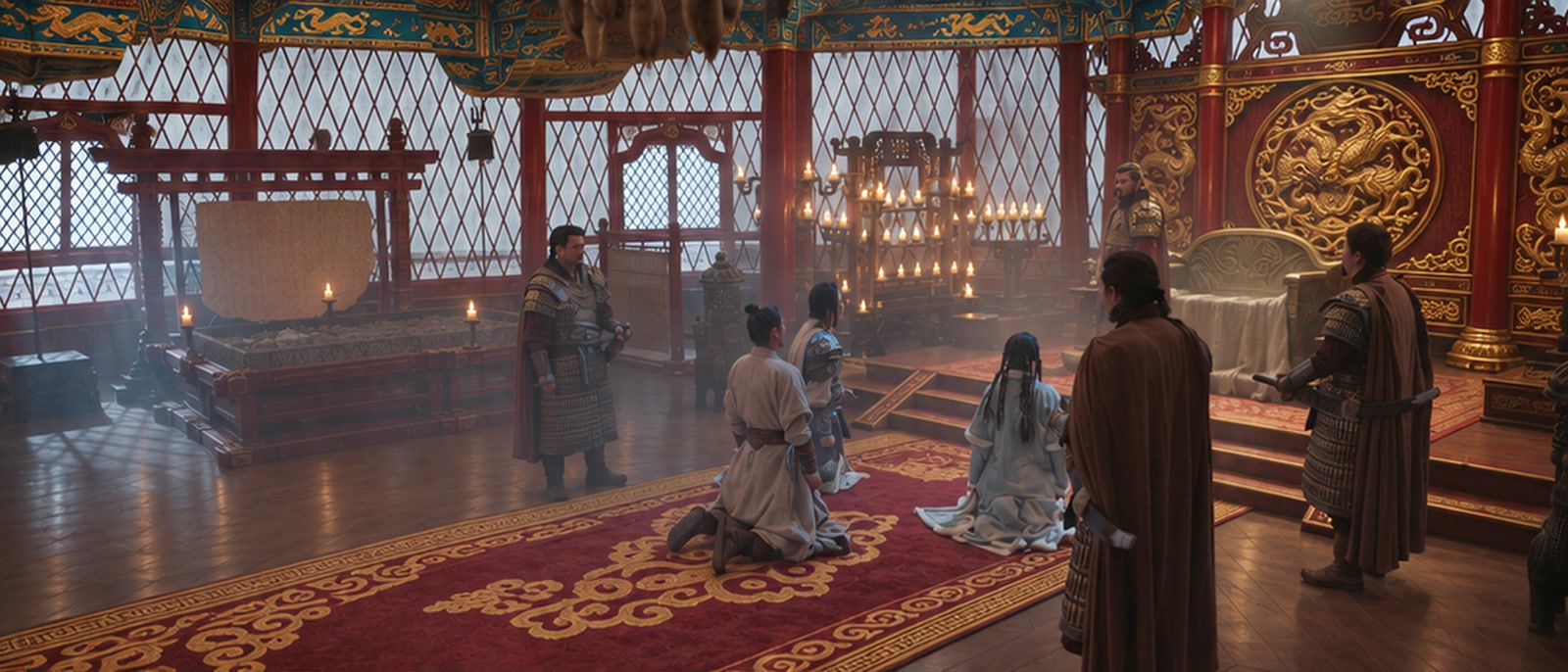}&
\includegraphics[width=.325\textwidth]{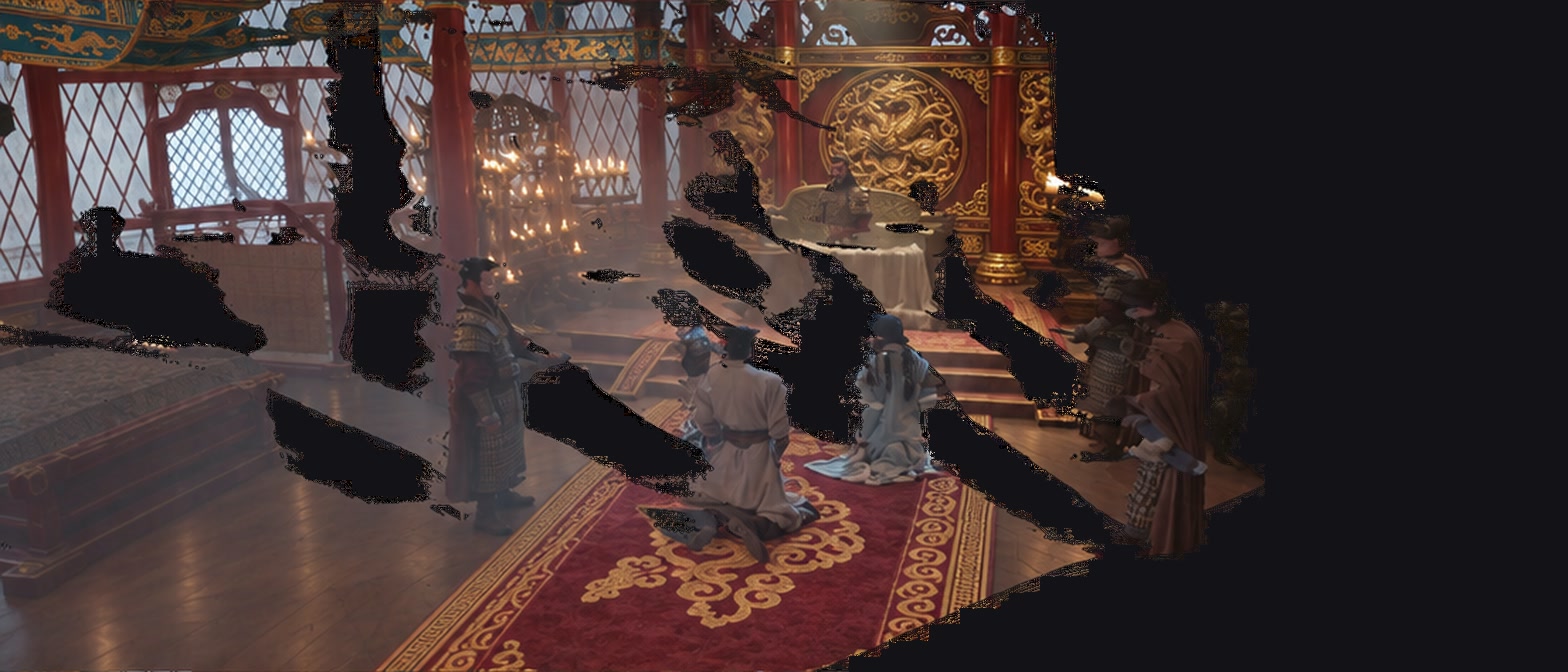}&
\includegraphics[width=.325\textwidth]{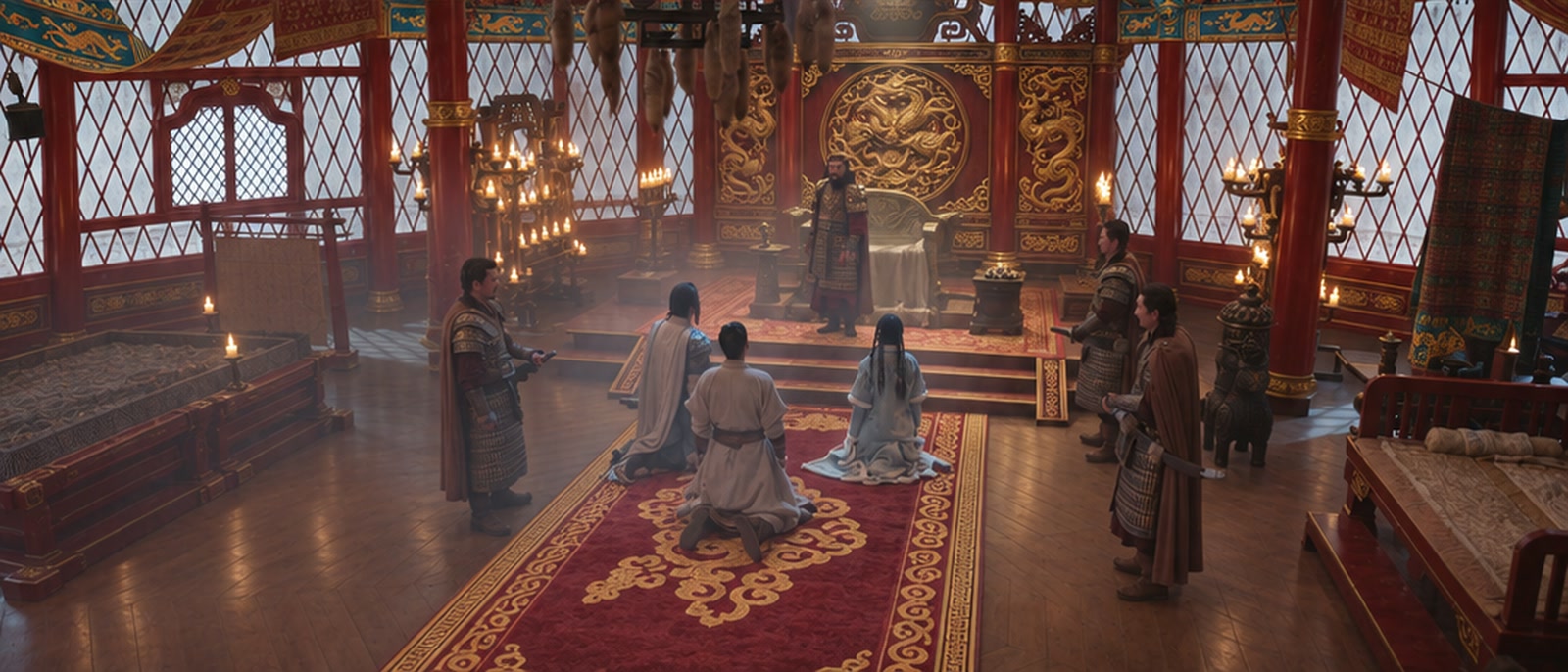}\\
\end{tabular}
\caption{\textbf{Real-world data.} Each row shows a source frame, the point-video frame rendered into the target camera from feed-forward reconstruction, and the target frame. The point videos are sparser than those of rendered data, with holes along depth discontinuities and with stretched surfaces, yet people and scene elements land where the target recording shows them. The images shown here are AI-edited for illustration.}
\label{fig:data-real}
\end{figure}

Each selected clip is then reconstructed from both cameras at its full frame rate, and its point videos are rendered in both directions.
The rendering parameters are chosen to remove outlier depth points. They also reduce the holes and flicker that limited point resolution and depth discontinuities cause.
Each clip thus yields two training triplets, $(V_A,G_{A\to B},V_B)$ and $(V_B,G_{B\to A},V_A)$. Figure~\ref{fig:data-real} shows examples.

\subsection{Reference images}
\label{sec:references}
When the camera moves from a character's back to their face, from a wide shot to a close-up, or toward background outside the source frame, part of the target is not determined by the source.
Character references supply faces, garment details, and the unseen side of the body.
Scene references supply background structure that becomes visible in the target.
Reference selection must avoid two failure modes.
A reference that repeats the source viewpoint adds little and tends to be ignored during training.
A reference too close to the target composition creates a copy-paste shortcut, so that the generator learns to copy the reference instead of reshooting the source.
Selection therefore balances novelty relative to the source against separation from the target.

We construct references in two ways.
When a take contains a third camera $C$ that shows the same character or scene, cameras $A$ and $B$ serve as source and target, and references are cropped from $C$. These references then show views that differ from both the source and the target.
When no suitable third camera exists, an image generation model produces references from the available observations, showing the character or scene at other shot scales, poses, or viewpoints.

\subsection{Structured caption}
\label{sec:caption}
A vision-language model (VLM) captioner reads the source video and the references and writes a structured caption.
The caption binds every reference to its character or scene, so that several images of one person share an identity description and the attributes of different people are not mixed.
It also distinguishes source content from artifacts of the point video.
The caption states that identity, clothing, action, and scene relationships must be retained, whereas holes, sparse texture, and reconstruction jitter in the point video must not be reproduced.

The caption deliberately contains no camera-motion instruction, such as ``pan left'' or ``dolly in'', and no quantitative description of camera angles.
Converting a continuous trajectory into words introduces ambiguity in direction, magnitude, and timing and creates a second camera signal that may conflict with the point video.
All camera conditioning is carried by the point video.
The captioner never sees the point video or the target recording, so the caption is identical for all target views of a source video.

\section{Method}
\label{sec:method}
\subsection{Overview}
\label{sec:task}
\begin{figure}[tbp]
\centering
\includegraphics[width=\textwidth,height=.52\textheight,keepaspectratio]{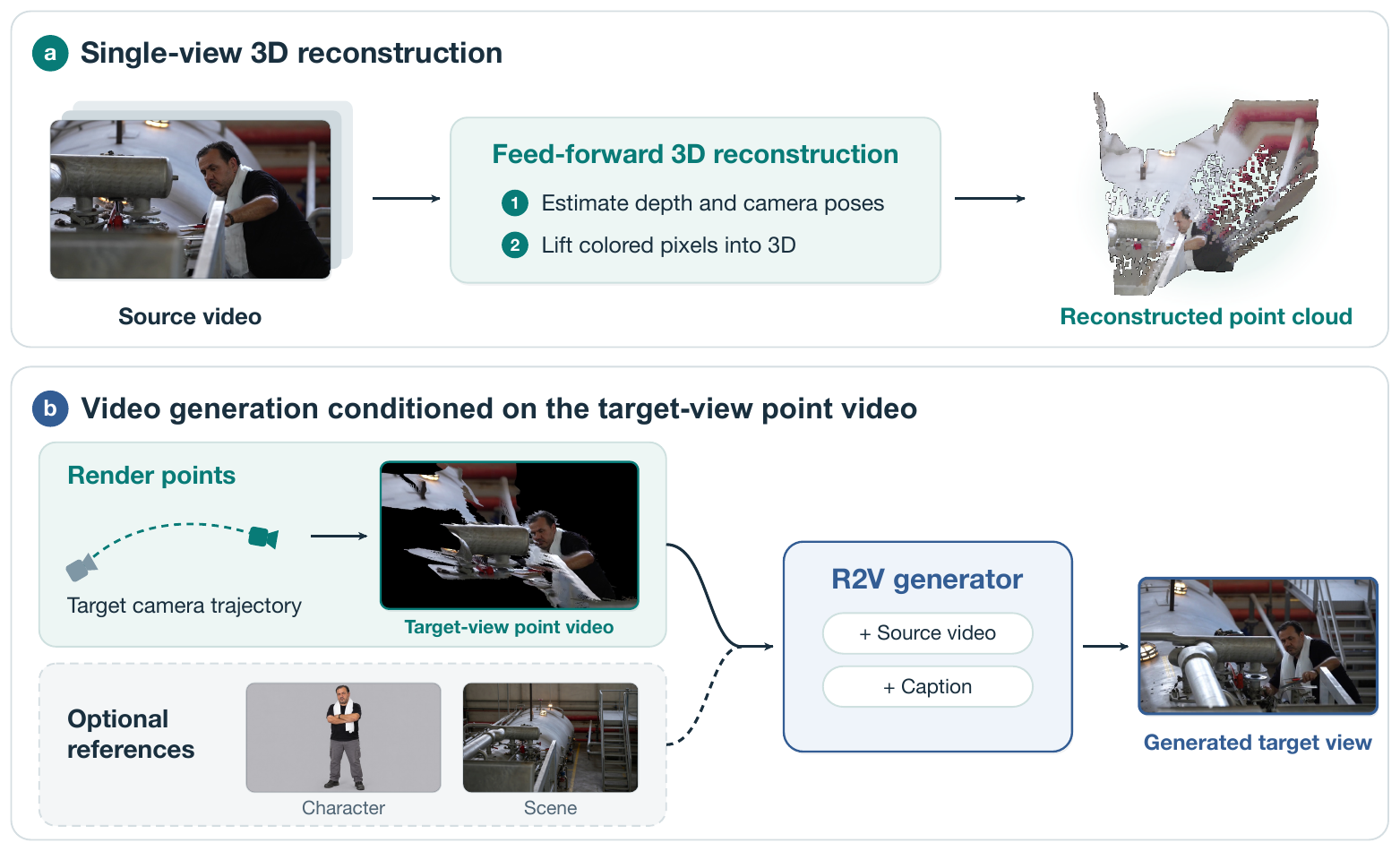}
\caption{\textbf{Video reshooting from a single source view.} (a) The system estimates per-frame depth and camera poses and lifts source pixels into a colored point cloud. (b) Rendering the points along the requested camera trajectory yields a target-view point video. The R2V generator uses this video, the source clip, a structured caption, and optional character and scene images to generate the new view.}
\label{fig:inference}
\end{figure}

Given a source video, one or more user-specified target camera trajectories, and optional character and scene reference images, \method{} produces one target video per trajectory on the timeline of the source (Figure~\ref{fig:inference}).
Each output keeps the subjects, clothing, scene relationships, and timing of the source.
Only the viewpoint and composition change.

A feed-forward reconstruction model estimates per-frame depth, intrinsics, and camera poses from the source video, and the VLM captioner of Section~\ref{sec:caption} writes the structured caption.
Both are computed once per source video and shared by all target views.
Training reconstructions use both cameras of a pair (Section~\ref{sec:data-real}), whereas inference uses the source video alone. We did not observe a difference between the two settings large enough to cause a domain gap.
For each target trajectory, the system renders the reconstruction along it into a point video (Section~\ref{sec:data-example}).
The system then encodes the source video, the point video, the references, and the caption and passes them to a pretrained R2V model adapted with low-rank adaptation (LoRA)~\citep{hu2022lora}, which generates the target video.
An optional super-resolution stage upscales the result (Section~\ref{sec:arch}).

\subsection{Architecture and super-resolution}
\label{sec:arch}
\label{sec:system}
\begin{figure}[tbp]
\centering
\includegraphics[width=\textwidth,height=.46\textheight,keepaspectratio]{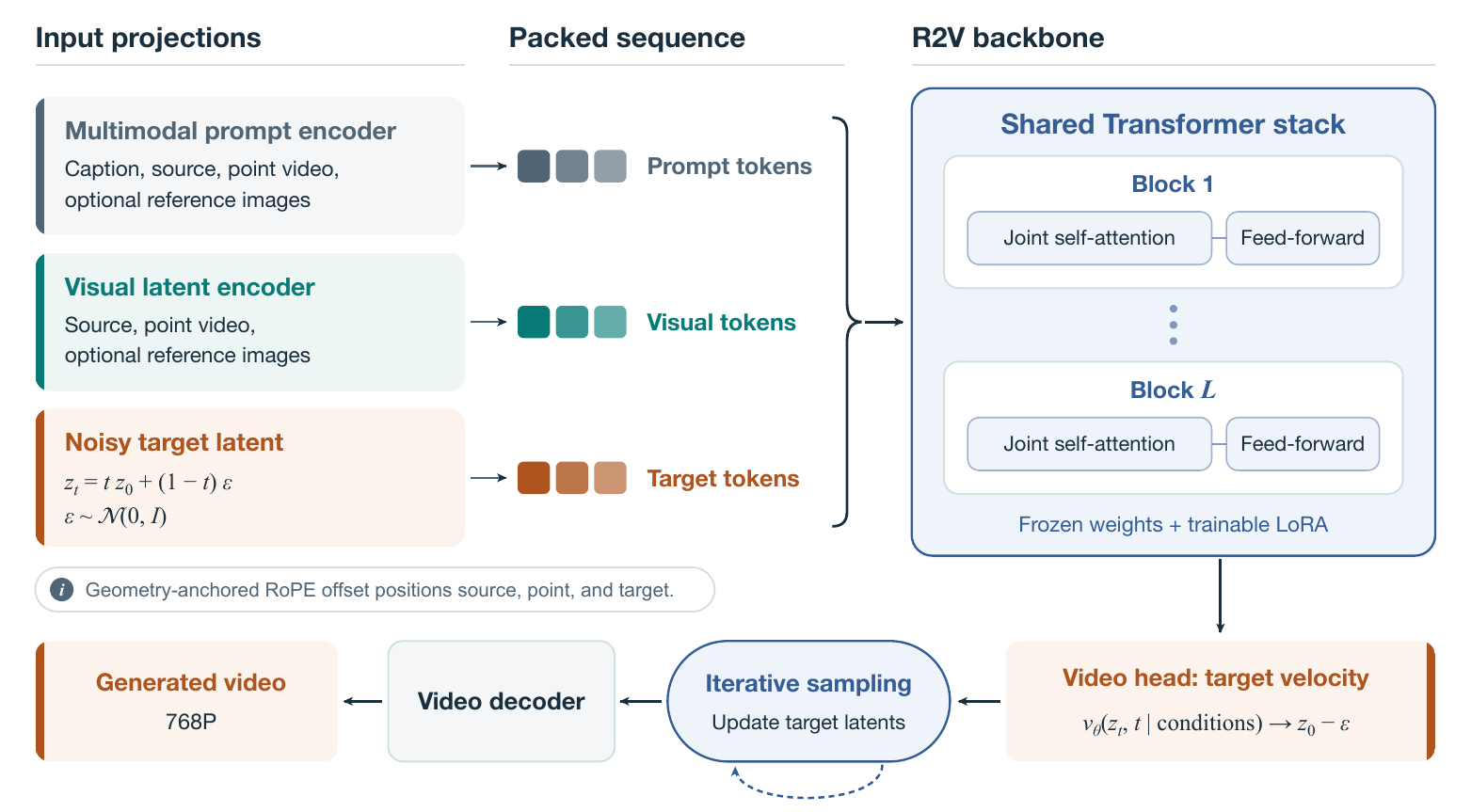}
\caption{\textbf{Conditional video generation with the R2V backbone.} Prompt, visual, and target tokens form one packed sequence processed by a stack of Transformer blocks with joint self-attention and trainable LoRA adapters. The video head predicts the target velocity for iterative latent updates. The decoder maps the final latent to a 768P video.}
\label{fig:architecture}
\end{figure}

A multimodal prompt encoder reads the source video, the point video, the references, and the structured caption.
A visual latent encoder maps the two videos and the reference images to latents.
Prompt features describe the subjects and what must be preserved, and visual latents retain appearance and structural detail.
The prompt features, the visual latents, and the noisy target latent are projected into prompt tokens, visual tokens, and target tokens, respectively, and concatenated into one packed sequence (Figure~\ref{fig:architecture}).
A single Transformer with joint self-attention processes the sequence, so target tokens can attend to all conditions at every layer.

\begin{figure}[tb]
\centering
\includegraphics[width=\textwidth]{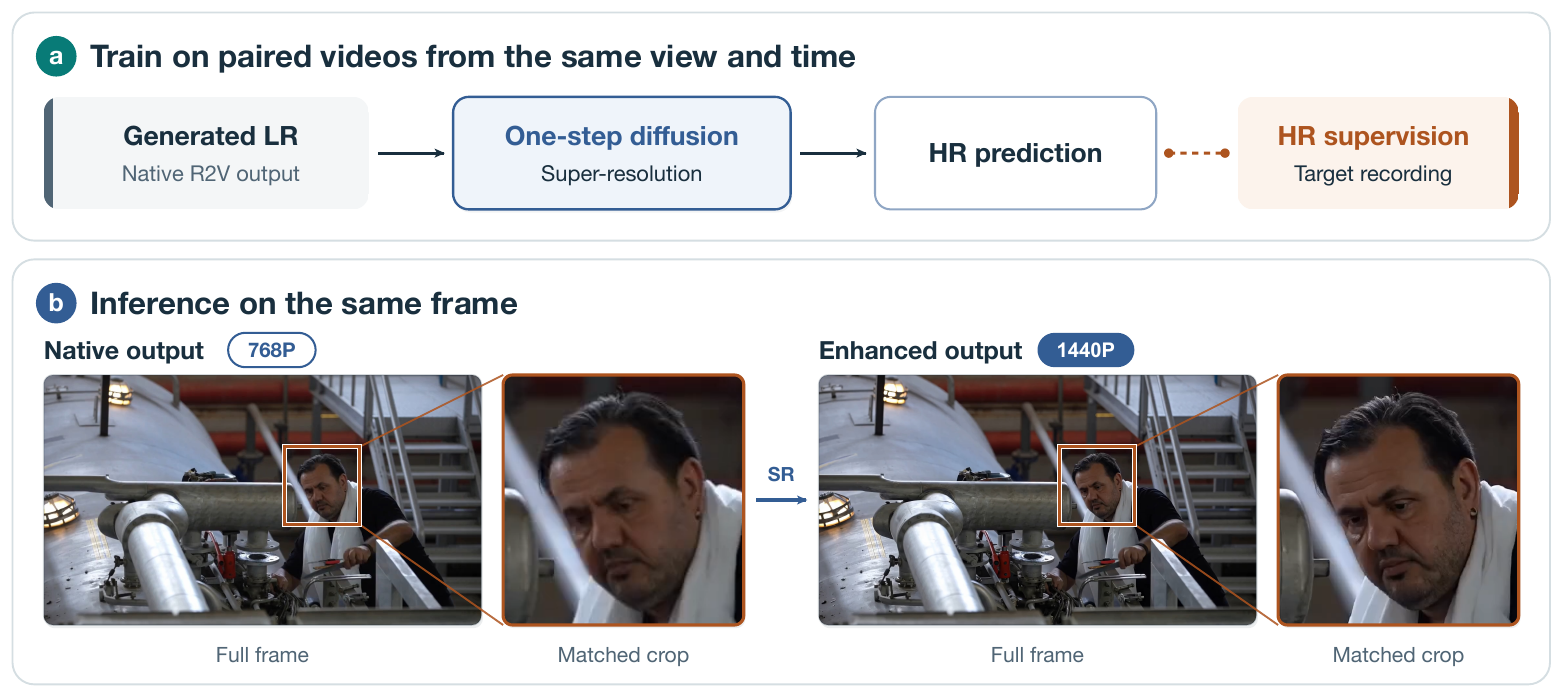}
\caption{\textbf{One-step video super-resolution.} (a) Schematic of a training pair in which the generated low-resolution video and the high-resolution target recording depict the same view at the same time. (b) A native 768P frame and its enhanced 1440P output are shown with crops of the same facial region.}
\label{fig:sr}
\end{figure}

\label{sec:sr}
The generator produces 768P video natively.
An optional one-step diffusion super-resolution stage upscales it to 1440P.
This stage is trained on pairs of generator outputs and the corresponding high-resolution target recordings, so it learns to restore the detail that the generator loses and to remove its artifacts, especially on faces, hair, and fabric (Figure~\ref{fig:sr}).

\subsection{Geometry-anchored RoPE offset}
\label{sec:layout}
Every visual token of the packed sequence, conditions included, carries rotary positional embedding (RoPE) coordinates $(t,h,w)$, and the attention between two tokens depends on the difference between their coordinates, so the coordinates act like addresses.
This has two consequences, which we call \emph{alignment} and \emph{copying}.
A target token tends to attend more strongly to condition tokens whose address is close to its own.
When a condition token has exactly the same address, its content tends to be copied into that target position.
Some training-free image-editing methods use copying on purpose.
FreeFlux identifies layers of RoPE-based diffusion Transformers whose attention is driven mainly by position and injects content through them~\citep{wei2025freeflux}.
CharaConsist moves reference tokens to the positions of corresponding points in the target so that character details are carried over~\citep{wang2025characonsist}.

\begin{figure}[tb]
\centering
\includegraphics[width=.86\textwidth]{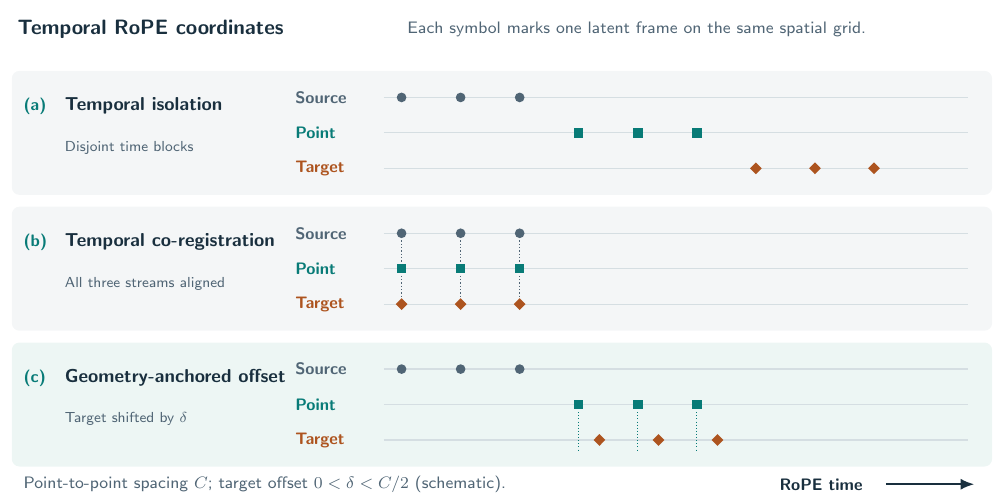}
\caption{\textbf{Temporal RoPE layouts.} Circles, squares, and diamonds mark source, point, and target frames, respectively. (a) Each stream occupies a separate temporal block. (b) Source, point, and target frames at each timestamp share the same RoPE coordinates. (c) The target coordinates shift by $0<\delta<C/2$ relative to the corresponding point frames, where $C$ is the spacing between adjacent point frames.}
\label{fig:layout}
\end{figure}

For the point video, we want alignment but not copying.
The output should follow the point video in time and space, so that actions stay synchronized with the source, and people and scene elements appear where the point video places them.
At the same time, the output must not copy the point video, whose appearance is degraded.
The source video raises only the copying concern. Its people and scene elements appear where the source camera sees them, so a target token that shared the address of a source token would paste them into the target view at the wrong place.
The addresses given to the two conditions relative to the target decide whether each effect occurs.
We consider three layouts (Figure~\ref{fig:layout}).
\begin{itemize}[leftmargin=1.2em]\setlength{\itemsep}{1pt}
\item \textbf{Temporal isolation.} Source, point, and target occupy consecutive, disjoint temporal blocks. This is how the backbone concatenates reference videos by default. Corresponding point and target tokens are a full clip length apart, so their temporal correspondence is weak. As a result, the actions in the output fall out of sync with those in the source and point videos.
\item \textbf{Temporal co-registration.} Source, point, and target share one temporal origin, so corresponding tokens of the three streams have identical coordinates. The output copies content from both conditions, so the same people can appear twice, once at their position in the source view and once at their position in the point video. The copies from the point video also inherit its degraded appearance, and image quality drops.
\item \textbf{Geometry-anchored offset} (used). The source keeps its own temporal block. Target tokens share the spatial grid of the point video and follow its temporal coordinates with a fixed offset of less than half a latent frame.
\end{itemize}

Let $j$ index latent frames, let $C$ be the RoPE temporal step between adjacent latent frames, let $a_s$ be the temporal origin of the source block, and let $L_s$ be its temporal span.
The geometry-anchored offset assigns
\begin{align}
 p_\mathrm{src}(j,h,w)&=(a_s+jC,\ h,\ w),\nonumber\\
 p_\mathrm{pt}(j,h,w)&=(a_s+L_s+jC,\ h,\ w),\label{eq:layout}\\
 p_\mathrm{tgt}(j,h,w)&=(a_s+L_s+jC+\delta,\ h,\ w),\qquad 0<\delta<C/2.\nonumber
\end{align}
Along the temporal axis, the source and the point video fill two consecutive blocks with one step $C$ per latent frame.
Each target frame sits a distance $\delta$ after the point-video frame of the same moment, on the same $(h,w)$ grid.
Because $\delta<C/2$, the point-video frame closest to target frame $j$ is frame $j$ itself, at distance $\delta$, rather than its neighbors at distance $C-\delta>C/2$ or $C+\delta$, so the output follows the timing of the point video.
Because $\delta>0$, no target token has exactly the same address as a point-video token.
Copying is therefore weaker, and the output takes its appearance mainly from the source rather than from the point video.
Section~\ref{sec:abl-layout} compares the three layouts.

\subsection{Training}
\label{sec:training}
\label{sec:stages}
\label{sec:data-mix}
\label{sec:objective}
The pretrained R2V model is kept frozen, and LoRA adapters are trained with the flow-matching objective~\citep{lipman2022flowmatching} of the base model.
Training proceeds in two stages that share this objective.

\paragraph{Stage 1: rendered data.}
The first stage trains on rendered data only.
Their point videos are clean and agree exactly with the target recordings, so the R2V model learns how a point video relates to the target camera and what the reshooting task asks it to preserve.
A model trained only on such data, however, learns to follow and largely copy reliable point videos (Section~\ref{sec:abl-real}).

\paragraph{Stage 2: rendered and real data.}
The second stage continues training on a mixture of real and rendered data at a ratio of 9:1 (real to rendered).
It resembles supervised fine-tuning on data close to the target application.
Real examples match the inference input in image appearance and in point-video quality (Table~\ref{tab:consistency}).
Their point videos show defects like those in Figure~\ref{fig:extreme}, and each of them is nevertheless paired with a correct target recording.
From these examples, the generator learns to take only the placement and orientation of people and scene elements from the point video, and their appearance from the source video and the references.
On real videos, this stage reduces the transfer of point-video artifacts into the output (Section~\ref{sec:abl-real}).

\section{Evaluation}
\label{sec:evaluation}
We evaluate whether \method{} produces usable reshot videos, with a focus on the property highlighted in Figure~\ref{fig:extreme}: following the requested camera and completing people and scenes when the point video is strongly degraded.

\subsection{Benchmark and settings}
\label{sec:setup}
\paragraph{Test data.}
For the quantitative and human evaluation in Tables~\ref{tab:geo} and~\ref{tab:gsb}, we use 50 cases from multi-camera takes that were not used in training.
Each case pairs a source video with a synchronized video recorded by the target camera.
To group the cases, we average over frames the fraction of target pixels covered by the point video.
The small-viewpoint-change group contains the 25 cases with coverage above 50\%, and the large-viewpoint-change group contains the 25 with coverage below 50\%.

\paragraph{Compared methods.}
For each case, every method generates a video from the target viewpoint.
Our method and Vista4D~\citep{lin2026vista4d} use the source video and the point video of the target trajectory.
CameraAnything~\citep{li2026cameraanything} takes the source video and the source-to-target relative camera pose as its camera control input.
The point video is rendered from the feed-forward reconstruction model of our system, which reconstructs these clips better than the default reconstruction of Vista4D and therefore does not disadvantage it.
Otherwise, we run both baselines with their public settings.
Our method receives no character or scene references in this comparison.
All methods are evaluated at their native output resolution without super-resolution.

\subsection{Objective evaluation}
\label{sec:quantitative}
We measure how closely each output follows the requested camera.
We apply the same feed-forward reconstruction model separately to (source, generated video) and (source, target).
At each synchronized frame, the first reconstruction estimates the generated camera pose relative to the source camera, and the second provides the corresponding reference pose estimate from the recorded views.
We compute pose errors between these estimates frame by frame.
RotErr is the mean geodesic angle between the generated and reference rotations, in degrees.

The two reconstructions share the source frames but have independent scales.
We estimate a single scale factor $s$ for the clip by taking the median of the depth ratios (reference over generated) at corresponding valid source pixels across all frames.
Let $p_t^{\mathrm{gen}}$ and $p_t^{\mathrm{ref}}$ be the two target-camera positions in source-camera coordinates at frame $t$. We apply only the scale $s$ to $p_t^{\mathrm{gen}}$ and normalize the mean position error over the $N$ frames by the median reference baseline $b$.
\[
b = \operatorname{median}_{t}\lVert p_t^{\mathrm{ref}}\rVert_2 ,
\qquad
\mathrm{TransErr} = \frac{1}{N}\sum_{t=1}^{N}
    \frac{\lVert s\,p_t^{\mathrm{gen}}-p_t^{\mathrm{ref}}\rVert_2}{b} .
\]
TransErr measures errors in both the direction and magnitude of camera displacement.
Table~\ref{tab:geo} reports mean RotErr and TransErr for each coverage group and for all 50 pairs, weighting each pair equally.

\begin{table}[tb]
\centering\small
\captionsetup{position=top,skip=18pt}
\caption{\textbf{Relative camera pose errors on synchronized video pairs.} Pose errors use the relative poses estimated from source and recorded target frames as the reference. TransErr is expressed as a fraction of the reference clip's median baseline.}
\label{tab:geo}
\begin{tabular}{@{}lrr@{}}
\toprule
Method & RotErr (deg) $\downarrow$ & TransErr $\downarrow$ \\
\midrule
\multicolumn{3}{@{}l}{\emph{All pairs ($n=50$)}} \\
Vista4D~\citep{lin2026vista4d} & 2.913 & 0.730 \\
CameraAnything~\citep{li2026cameraanything} & 3.067 & 0.925 \\
Ours & \textbf{1.895} & \textbf{0.564} \\
\midrule
\multicolumn{3}{@{}l}{\emph{Small viewpoint change: pixel coverage $>50\%$ ($n=25$)}} \\
Vista4D & \textbf{1.076} & 0.502 \\
CameraAnything & 2.193 & 0.961\\
Ours & 1.580 & \textbf{0.333} \\
\midrule
\multicolumn{3}{@{}l}{\emph{Large viewpoint change: pixel coverage $<50\%$ ($n=25$)}} \\
Vista4D & 4.751 & 0.957 \\
CameraAnything & 3.941 & 0.888 \\
Ours & \textbf{2.210} & \textbf{0.795} \\
\bottomrule
\end{tabular}
\end{table}

Table~\ref{tab:geo} shows that our method achieves the lowest overall RotErr and TransErr.
For small viewpoint changes, Vista4D attains a lower RotErr.
Vista4D depends more strongly on the point video and follows its positions closely. This helps while the point video is reliable.
When a large viewpoint change degrades the point video until it is hard to follow, the RotErr of Vista4D rises from 1.076 to 4.751 degrees, whereas ours rises only from 1.580 to 2.210.
Our method then ranks first on both metrics, which reflects the robustness to degraded point videos illustrated in Figure~\ref{fig:extreme}.

\subsection{Human evaluation}
\label{sec:human}
We compare the outputs through pairwise Good/Same/Bad (GSB) judgments.
Raters see the source video, the point video, which shows the requested camera, and two anonymized outputs in random order.
They judge whether the output of our method is better (Good), equivalent (Same), or worse (Bad) along the following dimensions.
\begin{itemize}\setlength{\itemsep}{1pt}
\item \textbf{Camera adherence.} Whether the output shows the scene from the requested viewpoint and follows the requested camera motion.
\item \textbf{Identity and appearance.} Whether performers keep the face, hairstyle, clothing, and accessories of the source video, and whether the scene keeps its layout and materials.
\item \textbf{Temporal coherence.} Whether actions stay synchronized with the source and whether the video is free of flicker, jumps, and drifting identity.
\item \textbf{Visual quality.} Whether the frames are sharp and free of artifacts such as distorted faces or limbs, streaks, holes, and garbled texture.
\item \textbf{Overall preference.} Which output the rater would rather use as the reshot video.
\end{itemize}
Table~\ref{tab:gsb} reports the net GSB score $(G-B)/(G+S+B)$ as a percentage.
Across all five dimensions, our method obtains positive net scores against both baselines, with overall scores of 38\% against Vista4D and 92\% against CameraAnything.

\begin{table}[tb]
\centering\small
\setlength{\tabcolsep}{4pt}
\captionsetup{position=top,skip=18pt}
\caption{\textbf{User preference (GSB).} Each cell reports the net GSB score $(G-B)/(G+S+B)$ as a percentage for our method against the baseline. Positive values indicate a preference for our method.}
\label{tab:gsb}
\begin{tabular}{@{}lccccc@{}}
\toprule
Comparison & Camera & Identity \& app. & Temporal & Quality & Overall \\
\midrule
Ours vs. Vista4D & 4\% & 18\% & 28\% & 22\% & 38\% \\
Ours vs. CameraAnything & 72\% & 68\% & 36\% & 68\% & 92\% \\
\bottomrule
\end{tabular}
\end{table}

\subsection{Qualitative comparison}
\label{sec:qualitative}
\begin{figure}[!t]
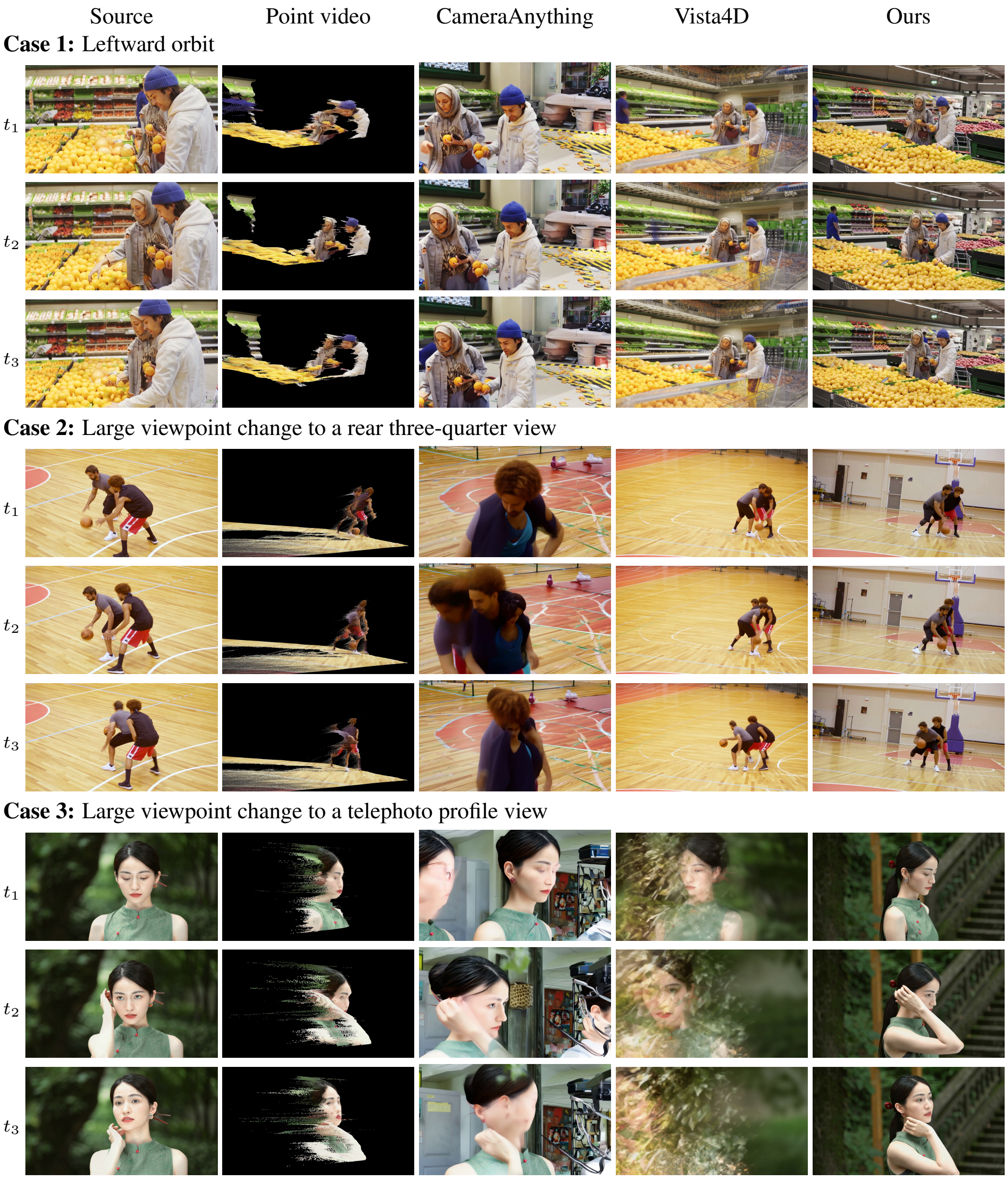

\centering
\small
\setlength{\tabcolsep}{1pt}\renewcommand{\arraystretch}{0.85}
\newcommand{\qualiimg}[3]{\includegraphics[width=.1895\linewidth]{figures/fig10_qualitative/quali_#1_#2_f#3.jpg}}
\newcommand{\qualirow}[2]{\raisebox{.043\linewidth}{\scriptsize $t_{#2}$} & \qualiimg{#1}{src}{#2} & \qualiimg{#1}{pc}{#2} & \qualiimg{#1}{camany}{#2} & \qualiimg{#1}{vista4d}{#2} & \qualiimg{#1}{ours}{#2}}
\begin{tabular}{@{}c@{\hspace{2pt}}ccccc@{}}
 &  Source &  Point video &  CameraAnything &  Vista4D &  Ours\\[2pt]
\multicolumn{6}{@{}l}{\textbf{Case 1:} Leftward orbit}\\[2pt]
\qualirow{1}{1}\\
\qualirow{1}{2}\\
\qualirow{1}{3}\\[2pt]
\multicolumn{6}{@{}l}{\textbf{Case 2:} Large viewpoint change to a rear three-quarter view}\\[2pt]
\qualirow{2}{1}\\
\qualirow{2}{2}\\
\qualirow{2}{3}\\[2pt]
\multicolumn{6}{@{}l}{\textbf{Case 3:} Large viewpoint change to a telephoto profile view}\\[2pt]
\qualirow{3}{1}\\
\qualirow{3}{2}\\
\qualirow{3}{3}\\
\end{tabular}
\caption{\textbf{Comparison with Vista4D and CameraAnything.} Each row shows the source, the point video, and the three generated views at the same timestamp. Frames run from top to bottom within each case. All methods use the same source clips and target trajectories.}
\label{fig:comparison}
\end{figure}

Figure~\ref{fig:comparison} shows three reshoots at matched timestamps.
In Case~1, our method places the shopper in blue behind the produce display even after the shopper leaves the source view. In the Vista4D sequence, the shopper disappears abruptly.
In Case~2, our method preserves the court layout and the relative positions of the two players as one dribbles past the other. Vista4D distorts their bodies and introduces what appears to be a third player at $t_3$.
In Case~3, a telephoto shot with shallow depth of field, our method follows the requested camera angle while preserving the woman's hand movement. Vista4D fails to follow the large viewpoint change, and the woman's face blends into the foliage.
Across these three cases, CameraAnything changes the framing or scene content enough to make its outputs largely unusable as reshoots (Table~\ref{tab:gsb}).
Appendix~\ref{app:results} presents additional qualitative examples.

\subsection{Analysis of design choices}
\label{sec:ablation}
Each analysis below varies one component of the full model and keeps the source clips and target trajectories fixed.

\begin{figure}[tb]
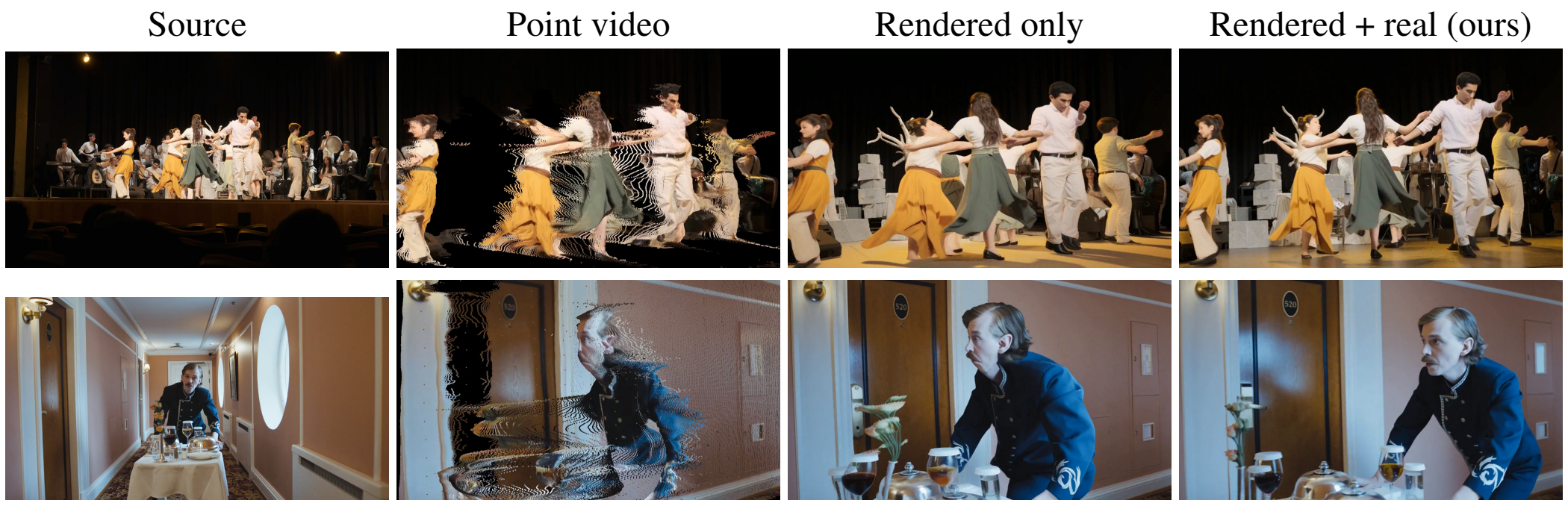

\centering
\setlength{\tabcolsep}{1pt}
\newcommand{\realablimg}[2]{\includegraphics[width=.243\linewidth]{figures/fig11_real_data_ablation/#1_#2.jpg}}
\newcommand{\realablrow}[1]{\realablimg{#1}{src} & \realablimg{#1}{pc} & \realablimg{#1}{WOreal} & \realablimg{#1}{ours}}
\begin{tabular}{@{}cccc@{}}
\small Source & \small Point video & \small Rendered only & \small Rendered + real (ours)\\
\realablrow{1}\\
\realablrow{2}
\end{tabular}
\caption{\textbf{Effect of real multi-camera training data.} Both variants reshoot the same source frame from the viewpoint of the point video. Real training footage reduces the transfer of point-video artifacts to the dancers (top) and to the attendant and the objects on the cart (bottom).}
\label{fig:abl-real}
\end{figure}

\paragraph{Real multi-camera data.}
\label{sec:abl-real}
We compare the full two-stage model of Section~\ref{sec:stages} with a variant trained only on rendered data in both stages.
The rendered-only model reproduces distortions visible in the point video (Figure~\ref{fig:abl-real}).
In the dance scene, facial features lose definition and the dancers' arms appear stretched or fused.
In the hallway, objects on the cart stretch sideways and the attendant's face is visibly deformed.
These distortions also jitter from frame to frame during playback (see the videos on the project page).
With real multi-camera training data, the full model better preserves the dancers' faces and poses and the shapes of objects on the cart, with less jitter across frames.

\begin{figure}[tb]
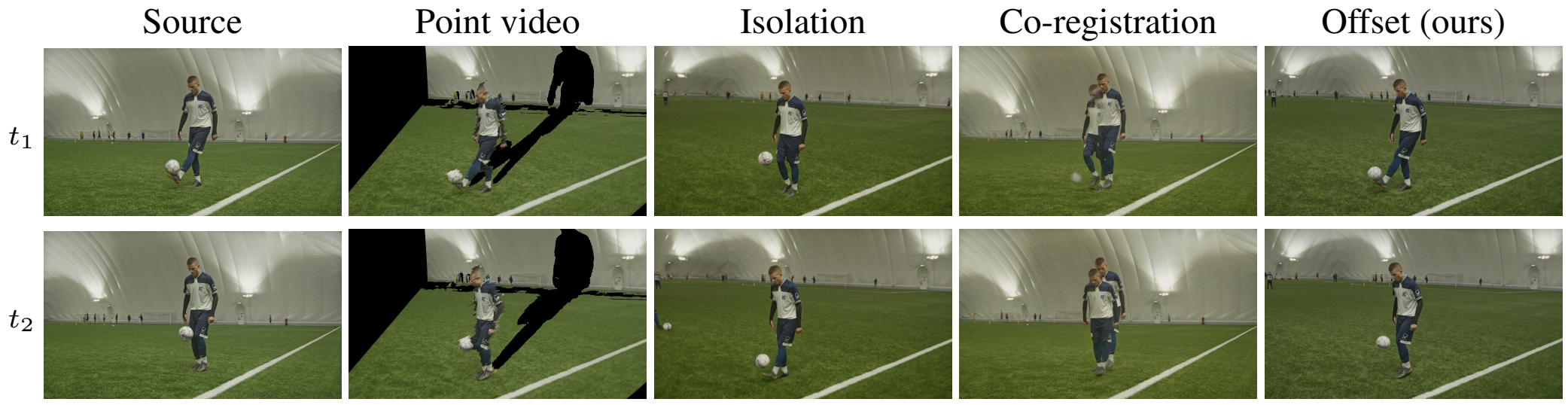

\centering
\setlength{\tabcolsep}{1pt}
\newcommand{\ropeimg}[2]{\includegraphics[width=.187\linewidth]{figures/fig12_rope_ablation/rope_1_#1_f#2.jpg}}
\newcommand{\roperow}[1]{\raisebox{.045\linewidth}{\scriptsize $t_{#1}$} & \ropeimg{src}{#1} & \ropeimg{pc}{#1} & \ropeimg{seq}{#1} & \ropeimg{co}{#1} & \ropeimg{ours}{#1}}
\begin{tabular}{@{}c@{\hspace{2pt}}ccccc@{}}
 & \small Source & \small Point video & \small Isolation & \small Co-registration & \small Offset (ours)\\
\roperow{1}\\
\roperow{2}
\end{tabular}
\caption{\textbf{Effect of the temporal RoPE layout.} Each row shows the same instant across views. Temporal isolation changes when the player's foot meets the ball. Co-registration superimposes two copies of the player. The geometry-anchored offset preserves both the timing of the touch and a single image of the player.}
\label{fig:abl-layout}
\end{figure}

\paragraph{RoPE offset.}
\label{sec:abl-layout}
We train one model with each of the three temporal layouts of Section~\ref{sec:layout} under otherwise identical settings and compare them on a player juggling a football (Figure~\ref{fig:abl-layout}).
With temporal co-registration, two overlapping, partly transparent copies of the player appear.
Temporal isolation produces a single player, but shifts the timing of the touch.
At $t_1$, the source shows the player's foot meeting the ball while the generated ball is already airborne.
At $t_2$, the ball has risen in the source but is back near the foot in the generated view.
With the geometry-anchored offset, the player appears once and the ball leaves the foot in step with the source.

\begin{figure}[!t]
\centering
\includegraphics[width=\linewidth]{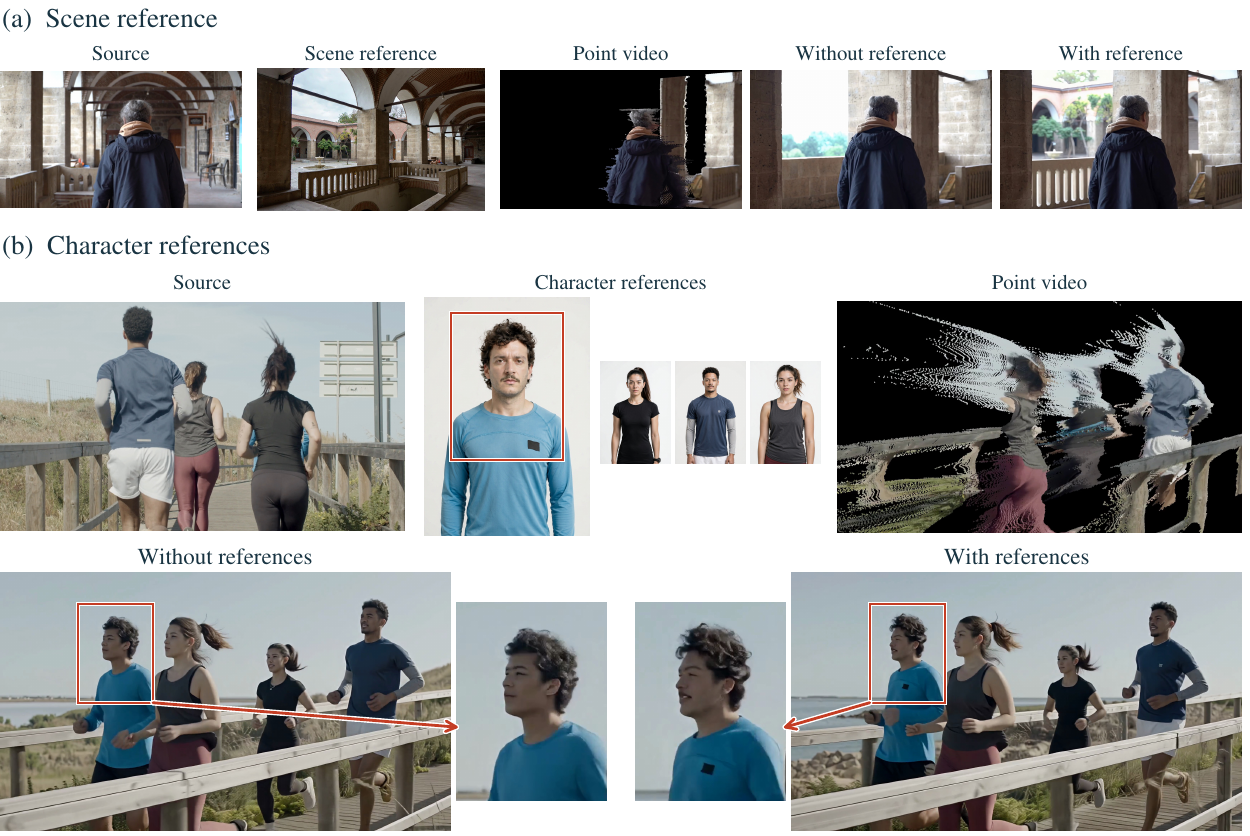}
\caption{\textbf{Effect of reference images.} (a) The scene reference supplies the appearance of the courtyard beyond the arcade. (b) The source shows the runners from behind. Boxes mark the face and upper chest of the runner in blue, in his reference portrait and in both target views. The enlargements show the shirt logo and its location relative to the collar.}
\label{fig:abl-ref}
\end{figure}

\paragraph{Reference images.}
\label{sec:abl-ref}
We generate each target view with and without reference images (Figure~\ref{fig:abl-ref}).
In the courtyard, the scene reference shows the arches across the open space. These arches appear behind the performer in the output with the reference, while the output without it leaves this region bright and largely featureless.
In the running sequence, the character reference shows a small dark logo on the chest of the runner in blue. The output with references reproduces the logo, while the corresponding area of the shirt is plain in the output without references.

\section{Limitations and Future Work}
\label{sec:limitations}
\paragraph{Limitations.}
\emph{Unobserved content and consistency across views.} Suppose the source camera films a performer from behind throughout a shot and the requested view shows their face. A frontal reference photo taken at another time can guide the shape of the face, but the model must still infer the performer's expression at each moment. The generated expression can look natural and still differ from the one a frontal camera would have recorded. Because the model generates each target view independently, two views can also depict different expressions at the same moment despite sharing the same reconstruction.


\emph{Additional performers.} In the multi-camera training footage, a target camera sometimes records a performer who is outside the source camera's field of view. Having learned from these pairs, the generator occasionally introduces an extra person when it reshoots a clip.

\emph{Identity drift in small faces.} When a face occupies only a few pixels in the native 768P output, the generator preserves little of the fine detail that distinguishes one performer from another. Super-resolution can then produce sharper facial features that resemble a different person.

\paragraph{Future work.}
\emph{Real-time streaming.} The current system processes prerecorded clips offline. Streaming would require estimating geometry from the frames received so far and updating it as new frames arrive. To keep pace with the input, we could pipeline reconstruction, rendering, and generation, reuse condition features, and test faster samplers.

\emph{Longer videos.} Generating a long shot in successive windows requires the model to preserve the same performers and scene across window boundaries. We plan to explore carrying subject and scene features forward and maintaining consistent temporal coordinates across windows to reduce gradual drift and abrupt changes at the boundaries.

\section{Ethical Considerations}
\label{sec:ethics}
The examples in this report and on the project page (\url{https://khuangvision.github.io/one_shot_any_view/}) are provided solely for technical demonstration and discussion. Film and television excerpts illustrate how the method behaves on cinematic footage. Rights to the original material remain with their respective holders.

A generated camera angle can show a face, expression, or gesture that the source camera never recorded. Even a plausible reconstruction can therefore change how an audience interprets a performance. Before using our system on footage of real people, users must obtain authorization from the people depicted for the proposed reshoot and its intended use, as well as permission to use the source footage and reference images. When sharing the results, users should identify the generated views so that viewers can distinguish them from recorded footage.

\section{Conclusion}
\method{} reshoots human-centric cinematic video from a single recorded shot.
Real-world multi-camera training data and our geometry-anchored RoPE enable it to generate high-quality reshot videos even when the point video is barely recognizable.
In both human evaluation and quantitative analysis, the system outperforms Vista4D and CameraAnything in human preference and camera accuracy.

\clearpage
{\raggedright
\bibliographystyle{plainnat}
\bibliography{references}
}

\clearpage
\appendix
\section{Additional Results}
\label{app:results}
\begin{figure}[!ht]
\centering
\includegraphics[width=\textwidth]{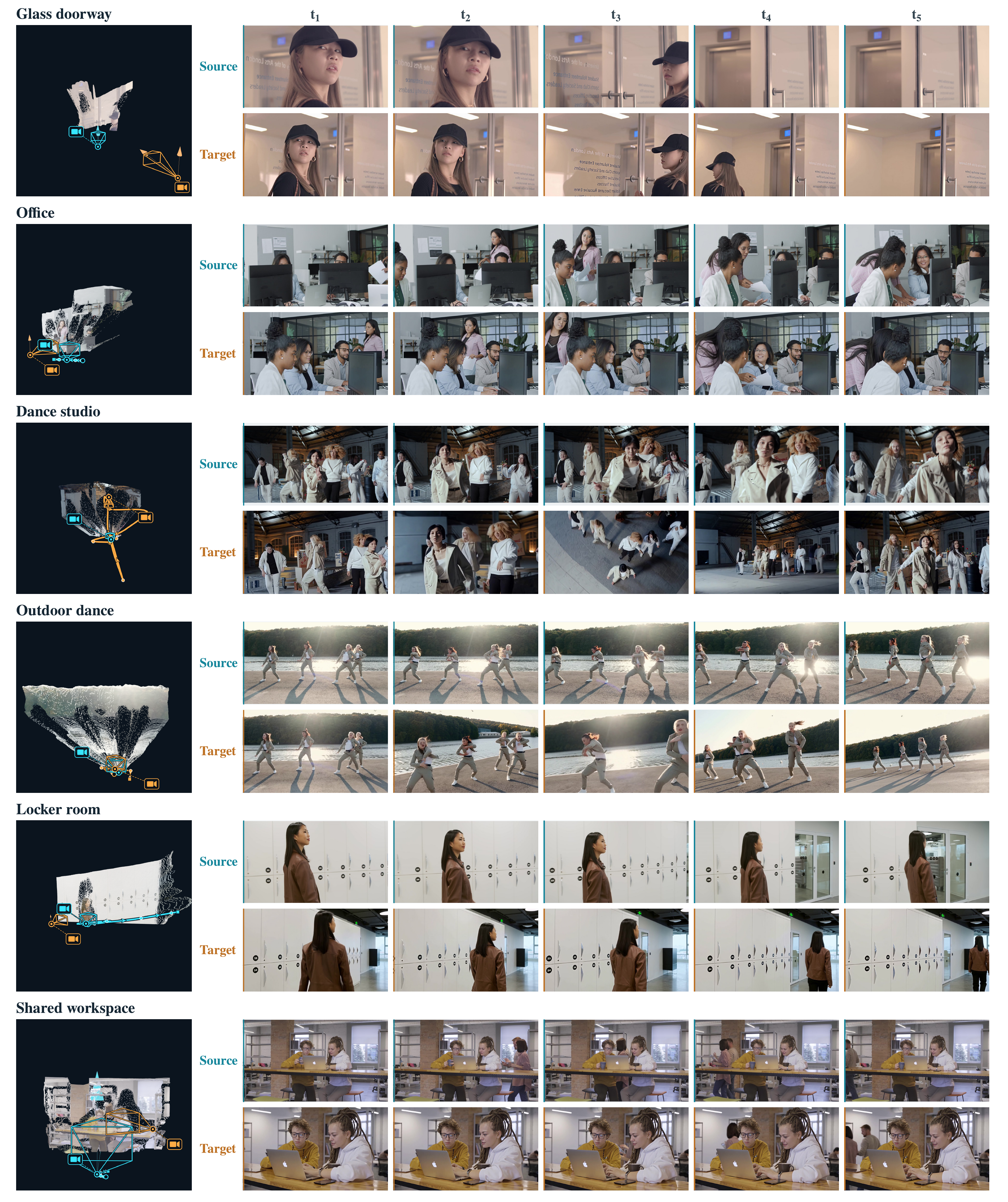}
\caption{\textbf{Additional reshooting results.} Each case shows the reconstructed point cloud and the source and target cameras on the left. On the right, source frames (top) and generated target-view frames (bottom) are paired at matching times, progressing from left to right. Cyan marks the source camera and orange marks the target camera.}
\label{fig:more-results}
\end{figure}

\paragraph{Additional reshooting examples.}
Figure~\ref{fig:more-results} shows six indoor and outdoor scenes. The examples include a glass doorway, an office, a dance studio, an outdoor dance, a locker room, and a shared workspace. The source and generated sequences show how the requested camera changes the framing while the performers continue their actions.

\end{document}